\documentclass[11pt]{article}
 \RequirePackage{fix-cm}
 
\usepackage{mathptmx}
\usepackage{mathrsfs}
\usepackage{amssymb, amsmath, amsfonts, mathtools}
\usepackage{amsthm}
\usepackage{epsfig}
\usepackage{graphicx}
\usepackage{subfigure}
\usepackage{epstopdf}
\usepackage{multicol}
\usepackage{array}
\usepackage{booktabs}

\usepackage{algorithm2e}
\usepackage {tikz}
\usetikzlibrary {positioning}
\usetikzlibrary{arrows}
\usetikzlibrary{decorations.pathreplacing}
\usepackage {xcolor}
\definecolor {processblue}{cmyk}{0.96,0,0,0}

\newcolumntype{C}[1]{>{\centering\let\newline\\\arraybackslash\hspace{0pt}}m{#1}}
\newcolumntype{L}[1]{>{\raggedright\let\newline\\\arraybackslash\hspace{0pt}}m{#1}}

\graphicspath{{images/}}

\begin{document}

\title{A Unified Framework for Nonmonotonic Reasoning with Vagueness and Uncertainty
}

\author{Sandip Paul\\
ECSU,Indian Statistical Institute, Kolkata
\and
Kumar Sankar Ray\\
ECSU,Indian Statistical Institute, Kolkata
\and
Diganta Saha\\
CSE Department\\
Jadavpur University}

\maketitle

\begin {abstract}

An interval-valued fuzzy answer set programming paradigm is proposed for nonmonotonic reasoning with vague and uncertain information. The set of sub-intervals of $[0,1]$ is considered as truth-space. The intervals are ordered using preorder-based truth and knowledge ordering. The preorder based ordering is an enhanced version of bilattice-based ordering. The system can represent and reason with prioritized rules, rules with exceptions. An iterative method for answer set computation is proposed. The sufficient conditions for termination of iterations are identified for a class of logic programs using the notion of difference equations.

  \end{abstract}
	

\section{Introduction:}

Answer set programming (ASP) \cite{gelfond1991classical,lifschitz1999answer,lifschitz2002answer} is a declarative problem solving paradigm for nonmonotonic reasoning. ASP allows intuitive representation of combinatorial search and optimization problems and it is widely used for knowledge representation and reasoning in various applications, like, plan generation, natural language processing etc \cite{eiter2009answer,esra2002theory}. Fuzzy answer set programming (FASP), \cite{saad2010disjunctive,saad2009extended,janssen2009general,van2007introduction,blondeel2014complexity,mushthofa2014finite,mushthofa2015solving}, possibilistic answer set programming \cite{nicolas2006possibilistic,bauters2012possibilistic,bauters2015characterizing} and probabilistic answer set programming \cite{baral2004probabilistic,de2013probabilistic} frameworks are extensions of classical ASP that can deal with imprecision and uncertainty. FASP allows graded truth values from the interval $[0,1]$ but cannot deal with uncertainty. Probabilistic and possibilistic approaches are constructed upon bivalent Boolean logic and hence are incapable of modeling the vagueness of information.

There are certain fuzzy logic programming frameworks that can represent and reason with uncertain knowledge. Some of the significant possibilistic logic-based approaches are: PLFC \cite{kullmann1999possibilistic,kullmann2001implementation}, a possibilistic logic for Horn clauses with fuzzy constants; PGL \cite{alsinet2000complete}, a possibilistic logic on top of Godel's infinite-valued logic; PGL$+$ \cite{alsinet2000complete1}, an extension of PGL for Horn Clauses and with fuzzy constants; P-DeLP,\cite{chesnevar2004logic} an argumentation theory based on PGL. Lukasiewicz \cite{lukasiewicz2009description} developed a probabilistic fuzzy description logic based on stratified fuzzy logic programs. A probabilistic logic programming approach defined over finite-valued Lukasiewicz logic is constructed \cite{lukasiewicz1999probabilistic}. Zhang et. al. developed a probabilistic fuzzy rule base by means of probabilistic fuzzy sets \cite{zhang2012efficient}, where a secondary probability distribution is imposed upon a primary membership function. But none of these approaches address answer set programming.

A possibilistic fuzzy answer set programming framework (PFASP), was proposed \cite{bauters2010towards} by merging possibilistic ASP with FASP. In PFASP two separate degrees, from the range $[0,1]$, are assigned to assert the degree of fuzziness and the degree of certainty of a proposition. But the proposed semantics is only for positive logic programs, without incorporating negation, neither classical negation nor negation-as-failure (not). Moreover, assigning two independent numbers as the degree of vagueness and the degree of uncertainty of a piece of information is not always intuitive. Because in human commonsense reasoning our assessment of the degree of truth of a proposition is somewhat dependent on the certainty about that proposition.

Interval-valued fuzzy sets (IVFSs) \cite{sambuc1975functions,deschrijver2007bilattice} represent vagueness and uncertainty simultaneously in an intuitive manner by replacing the crisp $[0,1]$-valued membership degree by sub-intervals of $[0,1]$. The intuition is that the actual membership would be a value within this interval. The intervals can be ordered with respect to their degree of truth as well as their degree of certainty by means of a bilattice-based algebraic structure, namely Bilattice-based Triangle \cite {arieli2004relating}. The truth and knowledge ordering play crucial role in determining the models of a logic program and their properties. A well-founded semantics of logic programs \cite{loyer2003approximate}is proposed, based on all sub-intervals of $[0,1]$ and bilattice-based ordering defined over them. Hybrid probabilistic logic programs \cite{dekhtyar2000hybrid} use intervals of probabilities and the truth and knowledge orderings of bilattice-baced triangle are interpreted as the degree of likelihood and degree of precision. Annotated answer set programming \cite{straccia2006annotated} uses interval-valued annotations as truth degrees of propositions and employs subset-based knowledge ordering for defining models of a program. Fuzzy Equilibrium logic \cite{schockaert2012fuzzy,schockaert2009answer}, an amalgamation of Pearce equilibrium logic\cite{pearce2006equilibrium} and FASP, also uses sub-intervals of $[0,1]$ and subset-based knowledge ordering over them.

However, the subset-based knowledge ordering in the bilattice-based triangle has a serious drawback when nonmonotonic reasoning under incomplete information is considered. It is demonstrated \cite{ray2018preorder} from the perspective of practical examples that; any proposition, which is somewhat true (i.e. assigned with an interval close to 1), can become false (say becomes $[0,0]$), when more information comes. These two valuations are not comparable with respect to the traditional knowledge ordering. This kind of knowledge comparison is necessary in multi-agent reasoning, where truth value of a proposition is assessed by different rational agents and the most certain assertion is to be taken as final. Belief revision in nonmonotonic reasoning also needs knowledge-comparability of two intervals when no one is a subset of the other. Preorder-based triangle is proposed \cite{ray2018preorder} as a modification of bilattice-based triangle. This algebraic structure employs preorders in place of lattice orders and can be thought as a fusion of Default bilattice \cite{ginsberg1988multivalued} and bilattice-based triangle.

This shift from the realm of lattice-based to preorder-based algebraic structures necessitates appropriate changes in the notion of satisfaction and definition of models in answer set programming. In bilattice-based approaches a preferred model is defined to be the greatest lower bound(glb) of the models; whereas for a preordered set no two elements have a unique glb. Application of preorder-based triangle for answer set semantics makes this work different from all the previous approaches.

An iterative approach for calculating the unified answer sets is proposed. The method is inspired from the standard computation of classical answer sets \cite{baral2003knowledge}. However, dealing with real numbers and product t-norm makes the analysis of convergence quite difficult. Here, we have formulated iterations in terms of difference equations and investigated the sufficient conditions for convergence by applying Contraction Mapping Theorem. The analysis procedure gives sufficient conditions for convergence of iterative computation for a broader class of programs than is considered in other FASP solving processes. For instance, FASP solvers are proposed for programs having simple loops with min t-norm \cite{janssen2012reducing}, for programs based on Lukasiewicz logic \cite{blondeel2014complexity} and for stratified positive logic programs only \cite{alviano2013fuzzy}.

The main contributions of this work are as follows:

$\bullet$ A modified version of Answer set Programming, named Unified Answer Set Programming (UnASP) is developed for nonmonotonic reasoning with vague and uncertain information. This uses all sub-intervals of $[0,1]$, ordered with Preorder-based triangle, as truth-space.

$\bullet$ An intuitive semantic definition of Negation-as-failure operator is given, which is helpful in computation of the answer sets.

$\bullet$ Unlike most of the proposed fuzzy answer set programming frameworks, that are constructed using min t-norm or Lukasiewicz t-norm, this approach uses product t-norm. It makes the approach more generalized since it is shown \cite{baaz1998embedding} that Godel and Lukasiewicz logics can be embedded in properly extended product logic.

$\bullet$ An interpretation is allowed to have both an atom and its negation, but having different levels of certainty. They stand for positive and negative evidences for a particular piece of information. A knowledge aggregation operator is defined, based on the modified knowledge ordering, that chooses the more certain epistemic state to include in the final answer set.

$\bullet$ An iterative approach, for computation of answer sets, is presented. This kind of iterative evaluation takes care of interval-valued interpretations for non-stratified programs. For the mathematical analysis of termination conditions, iterations are expressed by difference equations using the notion of computation graphs.

This paper is organized as follows: Section 2 describes the role of intervals in modeling uncertainty and vagueness and briefly describes the importance of preorder-based ordering of intervals over bilattice-based ordering. The logical connectives, considered in this work, are also described in this section. Section 3 describes UnASP rule structure and significance of various components. UnASP semantics is described in section 4, with proper illuminating examples. The iterative approach for answer set computation, mathematical analysis of iterations and several sufficient conditions for convergence is explained in section 5. Section 6 outlines a probable application domain. Proof of all theorems are presented together in the Appendix section.

\section{Intervals as degree of belief}

Selection of a proper set of truth-values, i.e., the set of values and their ordering is an important factor on which the appropriateness, performance and usability of an inference system depends. In absence of complete knowledge or in a multi-agent system, when different experts have different degrees of belief, ascribing an interval of possible values, taken from the unit interval, is the most intuitive solution \cite{nguyen1997interval}. An interval captures a rational agent's \textit{degree of belief} for a proposition. This epistemic interpretation of intervals may represent range of membership values for some fuzzy proposition or it can be interpreted as \textit{subjective probability} of a proposition. Intervals are appropriate to describe simultaneously the underlying imprecision and uncertainty of a piece of information.

\subsection {\textbf{Ordering of intervals based on degree of truth and certainty:}}

Let $\textbf{L} = \{L,\leq\}$, where $L$ is a complete lattice and the set of intervals of $L$ is defined as $I(L) = \{[x_1, x_2] | x_1, x_2 \in L $ and $ x_1 \leq x_2\}$. As a special case, the lattice $L$ is taken to be the unit interval $[0,1]$ and in that case $I([0,1])$ is the set of all sub-intervals of $[0,1]$, which stands for the truth space used in this approach. We will use the notation $\mathscr{T}$to denote $I([0,1])$ throughout this paper. 

The intervals can be ordered using their degree of vagueness (truth ordering, $\leq_t$) and their degree of certainty (knowledge ordering, $\leq_k$) by a bilattice-based triangle \cite{cornelis2007uncertainty} as follows:

\begin{center}

 $[x_1,x_2] \leq_t [y_1,y_2] \ \Leftrightarrow \ x_1 \leq y_1$ and $x_2\leq y_2$.

 $[x_1,x_2] \leq_k [y_1,y_2] \ \Leftrightarrow \ x_1 \leq y_1$ and $x_2\geq y_2$.

\end{center}
 
for every $[x_1,x_2]$ and $[y_1,y_2] \in \mathscr{T}$.

To demonstrate the shortcomings of bilattice-based knowledge ordering, we consider the prototypical example of assessing whether a bird, Tweety, can fly; given that it is not known whether Tweety is a penguin or not \cite{ray2018preorder}. In absence of complete information about Tweety, the degrees of belief of rational agents vary, say, from 0.7 to 1. Hence, the interval $[0.7,1]$ is assigned as epistemic state of the proposition "Tweety flies". Now suppose, some other information in the rule base (e.g., Tweety is a penguin or Tweety has broken wings) ensures that "Tweety cannot fly". This rule can be thought as another source of information in a multi-agent system. Following this, "Tweety flies" should get the epistemic state $[0,0]$. But in the bilattice-based triangle these two intervals are incomparable with respect to $\leq_k$. Though, it is evident that $[0,0]$ is an absolutely certain assignment and $[0.7,1]$ has some uncertainty, but the $\leq_k$ ordering is unable to reflect this. This type of situation is unavoidable in nonmonotonic reasoning, where rule bases have to be updated regularly. Moreover, The truth ordering $(\leq_t)$, used in bilattice-based triangle, does not order the overlapping intervals intuitively. Some more examples, depicting the short-coming of bilattice-based triangle are mentioned in \cite{ray2018preorder}. To salvage the problem some modifications were incurred to the knowledge and truth ordering \cite{ray2018preorder}.

For every $[x_1,x_2]$ and $[y_1,y_2] \in \mathscr{T}$ the modified truth and knowledge orderings are respectively as follows:

\begin{center}
 $[x_1,x_2] \leq_{t_p}[y_1,y_2]\Leftrightarrow x_m \leq y_m$.

 $[x_1,x_2] \leq_{k_p}[y_1,y_2]\Leftrightarrow x_w \geq y_w$.
\end{center}

where, for the interval $[x_1,x_2]$, $x_m$ is the midpoint of the interval $[x_1,x_2]$, given by $(x_1+x_2)/2$ and $x_w$ is the length of the interval, given by $(x_2-x_1)$. 

The algebraic structure that put together these two modified orderings is \textbf{preorder-based triangle} (\textbf{P}(L)), which is proved to be an enhanced version of bilattice-based triangle \cite{ray2018preorder}.

\begin{figure}
\begin{center}
\includegraphics [width=80mm]{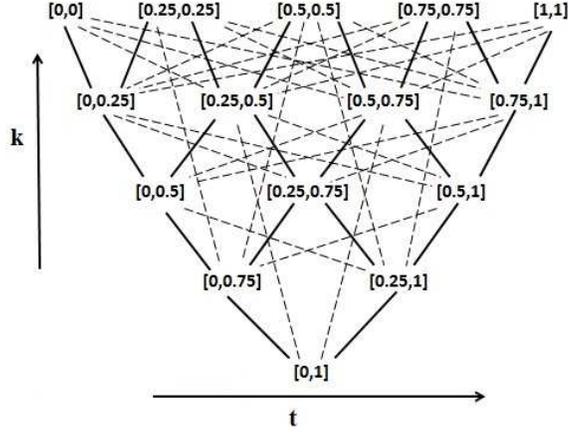}
\caption{Preorder-based Triangle for I(\{0,0.25,0.5,0.75,1\})}
\label{fig:largemoditri}
\end{center}
\end{figure}  

Figure \ref{fig:largemoditri} demonstrates all the intervals formed from the lattice $\textbf{L} =[\{0,0.25,0.5,0.75,1\},\leq]$ and ordered with $\leq_{t_p}$ and $\leq_{k_p}$. The dashed lines demonstrate the connections that were not imposed by $\leq_t$ and $\leq_k$. Thus from the figure it is clear that the modified orders introduce additional comparability between intervals. Exact intervals, i.e. intervals of the form $[x,x]$, may be used to characterize fuzzy and completely certain propositions. Assigning these intervals as epistemic states of propositions is as good as assigning a fuzzy truth value, i.e. a number from $[0,1]$. Intervals with non-zero length are used to express underlying uncertainty. Longer is the interval more is the uncertainty. The interval $[0,1]$ is used as the epistemic state of a proposition for which we have no knowledge at all, it can be true or false or something in between. 

The notation $x <_{t_p} y$ means $x \leq_{t_p} y$ and $x_m \neq y_m$; $x =_{t_p} y$ would be used if $x_m = y_m$. Similarly $x <_{k_p}$ and $x =_{k_p} y$ are interpreted. 

In this work, a preorder-based triangle is used to order the elements of over $\mathscr{T}$, based on which the modified answer set programming framework is developed. 

Though the enhanced comparability and mutual independence make the modified truth and knowledge ordering attractive, there is a well-developed theory of constructing conjunctors (t-norms) and disjunctors (t-conorms) for the traditional orders $\leq_t$ and $\leq_k$. Hence while building the unified answer set programming framework we will exploit both pairs of orders  appropriately to have the best of both worlds.

\subsection{Logical Operators}

The traditional logical operations defined over $[0,1]$ in Fuzzy Answer Set Programming like, classical negation, negation-as-failure, conjunctions, disjunctions can be generalized to define logical operations on $\mathscr{T}$. In this subsection the logical operators, used in this work, are discussed.

\subsubsection{Negation}

An involutive negator, namely the \textit{standard negator} (also known as Lukasiewicz Negator), $N$, which is a mapping $[0,1] \longrightarrow [0,1]: N(x) = 1-x$, can be used to define an involutive negator on $\mathscr{T}$ as follows:(\cite{cornelis2007uncertainty},\cite{ray2018preorder})

\begin{center}
$\textbf{N}([x_1,x_2]) = [N(x_2), N(x_1)]$

 $\ \ \ \ \ \ \ \ \ \ \ \  \ = [1-x_2, 1-x_1]$.

\end{center} 

Thus, it can be seen that strong negation doesn't alter the degree of certainty but reflects the interval around the central line of the preorder-based triangle. In the rest of the paper, standard negator will be represented using the symbol '$\neg$'.

\subsubsection{Negation-as-failure(not)}

In nonmonotonic reasoning, specially in logic programming, '\textit{not}' is used to draw inferences in absence of complete knowledge. The classical negation is different from \textit{'not'} in the way they deal with  incompleteness of information. In FASP, some monotonically decreasing function, or the Lukasiewicz negator is used to model negation-as-failure \cite{janssen2012core}. This function holds good there since FASP only deals with imprecision of information where a specific membership value can be given from $[0,1]$. However the situation is different when the uncertainty or lack of knowledge about the precise membership value is explicitly represented by using intervals of values. Here an intuitive alternative definition of negation-as-failure is developed as described below.

In nonmonotonic logic programming, for any atomic statement $p$, '\textit{not} $p$' is true if $p$ cannot be proved to be true or information about $p$ is absent. The statement $p$, being completely unknown, the epistemic state assigned to it is [0,1]. In absence of any information regarding statement $p$, \textit{not} $p$ is inferred to be true (i.e. the assigned epistemic state is [1,1]). Thus from an intuitive perspective, it can be said $not \ [0,1] = [1,1]$. Whereas, in case of standard negation $(\neg)$, we have $\neg [0,1] = [1-1,1-0] = [0,1]$; i.e, if no information is available about a statement then nothing can be said about its negation. 

The value of \textit{not} $p$ for an atomic statement $p$ would depend on how much knowledge about $p$ is available, as well as, on how much the experts believe in the truth of $p$. When $p$ represents absolutely certain but fuzzy attribute, i.e. has an exact interval as its epistemic state, then negation-as-failure would behave in the same way as classical negation. Information content, i.e. certainty level, of $\neg \ p$ is same as that of $p$. Thus if the interval is of the form $[x,x]$,
 
\begin{center}
$not [x,x] = \neg [x,x] = [1-x,1-x]$.
\end{center}

The interval assigned to $not \ p$ depends only on the epistemic state of $p$, which expresses the experts' degree of belief on $p$. Once the epistemic state of $p$ is obtained, $not \ p$ can be evaluated from it and hence the assignment for $not \ p$ does not directly come from experts' opinions, rather it's a meta level assignment based on the epistemic state of $p$, which is already asserted. Thus there would be no question of uncertainty while determining the epistemic state of $not \  p$, once information about $p$ is at hand. Therefore epistemic state of $not \ p$ will be exact intervals, solely depending on the epistemic state of $p$.

Hence, for an interval $[x_1,x_2] \in \mathscr{T}$

\begin{center}

$not [x_1,x_2] = [1-x_1, 1-x_1]$.

\end{center}

\begin{figure}
\begin{center}
\includegraphics [width=60mm]{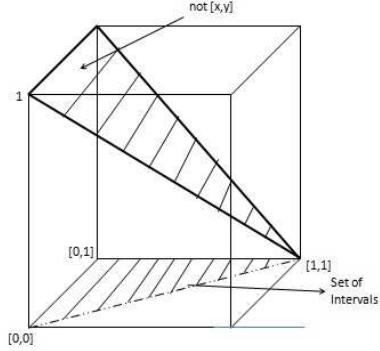}
\caption{3D plot of negation-as-failure}
\label{fig:not3d}
\end{center}
\end{figure}  

In Figure \ref{fig:not3d} the negation-as-failure is shown. From this definition it is clear that the operation '$not$' is not \textit{involutive} when applied on an interval with non-zero length and this is a significant difference between this definition and the way '$not$' is defined in FASP.

Here, we would like to point out that, negation-as-failure in annotated answer set programming framework \cite{straccia2006annotated}, gives rise to unintuitive result. The statement $not \ A:<0.2,0.3>$ means $I(A) \notin [0.2,0.3]$. Following this, we cannot express " when nothing is known about $A$" using nonmonotonic negation, because, $not \ A:<0,1>$ would be interpreted as $I(A) \notin [0,1]$, which is absurd, as it is evident that the truth degree of $A$ definitely lies in $[0,1]$. Whereas, the interpretation of $not$ provided here is intuitive and the purpose of $not$ in reasoning is served perfectly.

\subsubsection{Conjunctors and disjunctors}

T-representable t-norms and t-conorms are used as conjuctors and disjunctors for the set of intervals.

\textbf{Definition 1:}
A t-norm \textbf{T} on $\mathscr{T}$ is called t-representable if there exist t-norms $T_1$ and $T_2$ on $[0,1]$, such that $T_1 \leq T_2$ and such that \textbf{T} can be represented as, for all $[x_1,x_2],[y_1,y_2] \in \mathscr{T}$:

\begin{center}
\textbf{T}$([x_1,x_2],[y_1,y_2]) = [T_1(x_1,y_1), T_2(x_2,y_2)].$
\end{center} 
$T_1$ and $T_2$ are called representants of \textbf{T}. In this particular case $T_1$ and $T_2$ are identical.

Similarly a t-representable t-conorm or s-norms can be defined on $\mathscr{T}$ using two t-conorms on $[0,1]$.

In this work, the Product t-(co)norm is chosen as representant for constructing the t-representable t-(co)norm on $\mathscr{T}$. For two intervals $[x_1,x_2], [y_1, y_2] \in \mathscr{T}$, their t-(co)norm $\wedge$ and $\vee$ are defined respectively as follows:

   $[x_1,x_2] \wedge [y_1,y_2] = [x_1.y_1, x_2.y_2]$

	 $[x_1,x_2] \vee [y_1,y_2] = [x_1 + y_1 - x_1.y_1, x_2 + y_2 - x_2.y_2]$

\section{Syntax}

The constructed UnASP language has infinitely many variables, finitely many constants and predicate symbols, including comparative predicates like equality, less-than, greater-than etc. No function symbol is allowed. A term is a variable or a constant. An atom is an expression of the form $p(t_1, t_2,..,t_n)$, where p is a predicate symbol of arity n and $t_1, t_2,..,t_n$ are terms or elements of $\mathscr{T}$. An atom is \textit {grounded} if it contains no variables. A literal is a positive atom $a$ or its negation of the form $\neg a$. For a literal $l$, $not \ l$ is called a naf-literal.

A rule is of the form:

\begin{center}
$r: a \stackrel{\alpha_r}{\leftarrow} b_1, b_2, ... , b_m, not \ b_{m+1},  ... , not \ b_n$
\end{center}

where, $a$ and $b_i$ are positive or negated literals and $b_i$ can be elements of $\mathscr{T}$ too. The literal $a$ is the \textit{head} of rule. Here the expression $b_1, b_2,...,b_m, not \ b_{m+1}, .. , not \ b_n$ is the \textit{body} of the rule and it is actually a shorthand expression for $b_1 \wedge b_2 \wedge...\wedge b_m \wedge not \ b_{m+1} \wedge .. \wedge not \ b_n$, i.e.  the conjunction of literals. The label 'r' is used for this rule. The weight of the rule, i.e., the degree of truth and certainty of the rule, is specified by $\alpha_r$, which is an interval from $\mathscr{T}$. In other words, the weight denotes what would be the epistemic state of the consequent (or head) of the rule when the antecedent is absolutely true, i.e. assigned with $[1,1]$. The epistemic state (or truth status) of the body of a rule is combined with the weight by means of the product t-norm $(\wedge)$ and then it propagates to the head.
 
All rules are taken to be universally quantified and hence the quantification $(\forall x)$ is not specified explicitly. A rule is a \textbf{fact} if $b_i, 1\leq i \leq n$ are elements of $\mathscr{T}$.

A \textbf{Unified Answer Set Program} (UnASP) is a set of weighted rules as $r$. 

A program is called \textbf{positive} if for all rules in the program $n=m$, i.e. no naf literal is present in rule body and $a, b_1, b_2,...,b_m$ are positive atoms. A program is said to be \textbf{general} or \textbf{normal} if for every rule $a, b_1, b_2,...,b_n$ are positive atoms, i.e. don't contain classical negation $\neg$. Programs where naf literals as well as negated literals are allowed in the rules are syntactically referred to as \textbf{extended} programs.

For a program $P$, $P_l$ denotes the subset of $P$ consisting of the rules in $P$ whose head is the literal $l$.

\textbf{Significance of the rule weight $\alpha_r$ :}

The unified reasoning approach presented here is aimed to be a generalised framework suitable for reasoning with classical, vague as well as uncertain information. All these aspects can be captured by appropriately choosing the weights of the rules from $\mathscr{T}$.

$\bullet$ To represent statements that are free from any uncertainty and vagueness the weights are chosen to be $[1,1]$. For instance 'Birds gives egg' can be expressed as: 

\begin{center}
$r: \ GivesEgg(x)  \ \stackrel{[1,1]}{\leftarrow} Bird(x)$.
\end{center}

This rule states that, if '$x$ is a Bird' is True, i.e. assigned $[1,1]$, then '$x$ gives egg' is also True and is assigned the interval $[1,1]$.

$\bullet$ The weight $\alpha_r$ can be used to capture the \textit{vagueness}, where even if the body of the rule is satisfied the head may be partially true, having a degree of truth between $[0,0]$ and $[1,1]$. The proposition 'Small cars are moderately safe' can be formally written as:

\begin{center}
$r: \ Safe(x)  \ \stackrel{[0.7,0.7]}{\leftarrow} SmallCar(x)$.
\end{center}

Here, the weight $\alpha$ being an exact interval, $[0.7,0.7]$, depicts that the rule is certain but vague, i.e. the degree of safety varies over a range of [0,1].

$\bullet$ When statements from different sources are considered, they may have various degrees of reliabilities. Thus, wider intervals, as weights would imply lesser importance or reliability of the rule.

$\bullet$ The weight $\alpha_r$ helps to represent default statements which are assumed to be true under normal conditions but have exceptional cases. Default statements or \textit{dispositions} \cite{zadeh1985syllogistic} are used to represent commonsense knowledge or the commonplace statement of facts. Some examples are:

i.	Birds can fly.

ii.	Glue is sticky.

iii.Glass is fragile.

iv.Where there is smoke there is fire.

v.	Swedes are taller than Italians.

In a default rule, the conclusion is a plausible inference, drawn in absence of complete information. Hence even if the body of the default rule is satisfied the conclusion is not certain and can be inferred with some level of uncertainty. In this scenario, when the body is true, the weight of the rule will become the epistemic state of the head and hence play the role of a marker, designating that the corresponding information is attained in presence of some uncertainty and is subject to change when more concrete evidence comes into account. For instance, in this framework, the prototypical example of nonmonotonic reasoning concerning flying capability of birds can be modeled as follows:

$r1: \ Fly(x)  \ \stackrel{[0.7,1]}{\leftarrow} Bird(x), not Penguin(x)$

$r2: \ \neg Fly(x)  \stackrel{[1,1]}{\leftarrow} Penguin(x)$

$f1: \ Bird(Tweety) \stackrel{[1,1]}{\leftarrow}$.

Now, when performing reasoning about flying ability of Tweety, it can be seen that we have no information about whether Tweety is a Penguin or not. Hence, $not$ Penguin(Tweety) is True, i.e. $[1,1]$. From rule r1, $Fly(Tweety)$ is ascribed $[0.7,1]$. Here lies the significant difference between using weighted rules and un-weighted rules as in classical ASP. In classical ASP, $Fly(Tweety)$ would become True and hence bearing no trace that the information has some underlying uncertainty because of having incomplete information about Tweety. On the other hand, in case of UnASP, the epistemic state $[0.7,1]$ attached to $Fly(Tweety)$ will signify that it has some uncertainty.

$\bullet$ Another advantage of using weights, though not explored in this work, is variable weights, of the form $[\alpha,1]$, can be attached to dispositions, such that the value $\alpha$ varies with the number of exceptions and thus providing additional control and flexibility of knowledge representation.

\begin{center}
$Fly(x) \stackrel{[\alpha,1]}{\leftarrow} Bird(x)$
\end{center}

In general, as the uncertainty regarding the default rule increases, the weight $\alpha_r$ becomes a wider interval. Thus, using intervals of different widths, default rules can also be prioritized.

\section{Declarative Semantics}

The Atom Base $(\textbf{B}_P)$ for a program $P$ is the set of all grounded atoms and $\textit{Lit}_P$ is the set $\textit{Lit}_P = \{a | a \in \textbf{B}_P\} \cup \{\neg a | a \in \textbf{B}_P\}$. When no specific program is in the context of discussion then only $\textbf{B}$ and $\textit{Lit}$ are used.

For a UnASP program an \textbf{interpretation} is a mapping from the set of grounded literals to the set of intervals $\mathscr{T}$. An interpretation $\Im$ can be thought as a set of pairs $\Im = \{a: \bar{c} | a \in \Im^{Lit} \subseteq \textit{Lit}_P$ and $\bar{c} \in \mathscr{T} \}$. Cardinality of an interpretation, $\Im$, denoted by $|\Im|$, is the number of literals specified in $\Im$. An interpretation is \textbf{partial} if $|\Im| < |\textit{Lit}_P|$, i.e., $\Im^{Lit} \subset Lit_P$. Otherwise, it is said to be a \textbf{total} interpretation. If $a:[x_1,x_2] \in \Im$, then $\Im(a) = [x_1,x_2]$. 

Given an interpretation $\Im$, the epistemic states of all grounded atomic and complex expressions can be defined recursively as:

$v_{\Im}(\bar{c}) = \bar{c}$ for any $\bar{c} \in \mathscr{T}$;

$v_{\Im}(a) = \Im(a) = \bar{c}$ for $a:\bar{c} \in \Im$;

$v_{\Im}(a \wedge b) = v_{\Im}(a) \wedge v_{\Im}(b)$;

$v_{\Im}(a \vee b) = v_{\Im}(a) \vee v_{\Im}(b)$;

$v_{\Im}(not \ a) = not \ \Im(a)$.

where, $a,b \in \Im^{Lit} \cup \mathscr{T}$ or $a,b$ are complex expressions constructed from grounded literals using $\wedge, \vee, \neg$ and $not$. Here, we abused the notations $\wedge, \vee, \neg, not$ to mean the conjunction, disjunction and negation of literals and also to designate the mathematical operations to compute the epistemic states.

\textbf{Definition 2:} An interpretation is said to be consistent if for every pair of complementary literals $a:[x_1,x_2] \in \Im$ and $\neg a:[y_1,y_2] \in \Im$, 

i) $(x_2 - x_1) \neq (y_2 - y_1)$ or 

ii) $(x_2 - x_1) = (y_2 - y_1)$ and $x_1 + y_2 = 1$. In this case the corresponding interpretation is called \textit{strictly consistent} interpretation. 

Thus, for a strictly consistent interpretation $\Im$, if $\Im(a)  = [x_1,x_2]$ then $\Im(\neg a) = [1-x_2, 1-x_1]$. Therefore it is sufficient to mention the truth status of positive atoms from $\textbf{B}_P$ and values for the corresponding negated literals can be obtained from them; i.e. a strictly consistent total interpretation can be uniquely specified by its positive subset, that is the partial interpretation containing the positive atoms.

For any consistent interpretation, which is not strictly consistent, any pair of literals $a,\neg a$ are assigned epistemic states with different certainty degrees. In such a scenario $a$ and $\neg a$ are supposed to be completely independent of each other, i.e. two separate piece of information coming from different sources. But since both assertions are concerned about $a$, whether in positive or negative way, the more certain assignment is selected as the final epistemic state of $a$ for model construction.

The knowledge $(\leq{k_p})$ ordering, that are defined over the set of intervals, can also be imposed on the set of interpretations to order them.

For two interpretations $\Im$ and $\Im^*$, $\Im \leq_{k_p} \Im^*$ iff $\forall a \in Lit, \Im(a) \leq_{k_p} \Im^*(a)$.

\textbf{Definition 3:}

An interpretation $\Im$ is said to be \textit{k-minimal} of a set of interpretations $\Gamma$ if there is no other interpretation $\Im^* \in \Gamma$ such that $\Im^* \leq_{k_p} \Im$. If the \textit{k-minimal} interpretation is unique then it is called the \textit{k-least} interpretation of $\Gamma$.

\subsection{Satisfaction and Model:}

\textbf{Definition 4:}

Let P be any program, $\rho$ be any rule in P and $\Im$ be an interpretation of P. Let $r: a \stackrel{\alpha_r}{\leftarrow} B$ be a ground instance of $\rho$, then;

1. $\Im$ satisfies $r$ iff i) $v_{\Im}(a) = (v_{\Im}(B) \wedge \alpha_r)$ or $v_{\Im}(a) >_{k_p} (v_{\Im}(B) \wedge \alpha_r)$ or $v_{\Im}(a) >_{t_P} (v_{\Im}(B) \wedge \alpha_r)$.

2. $\models_{\Im} \rho$ iff $\Im$ satisfies every ground instance of $\rho$.

3. $\Im$ satisfies $P$, $\models_{\Im} P$, i.e. $\Im$ is a \textit{model} of P, iff $\Im$ satisfies every rule in P.

Hence, in simple words, a grounded rule is satisfied by an interpretation if the epistemic state of the head is equal to or strictly truer or strictly more certain that the product of the epistemic state of the body and the weight of the rule.

A logic program may have several models; among which a unique model or a set of models are chosen as the preferred models.

\textbf{Definition 5:}

A model $\Im_P$ of a program P is called \textbf{supported} iff;

 \  \  \  \  \   \  \  \  i. For every grounded rule $r: a \stackrel{\alpha_r}{\leftarrow} B$, such that $a$ doesn't occur in the head of any other rule, $v_{\Im_P}(a) = v_{\Im_P}(B) \wedge \alpha_r$.

 \  \  \  \  \  \  \  \  ii. For grounded rules $\{ a \stackrel{\alpha_1}{\leftarrow} B_1, a$     $\stackrel{\alpha_2}{\leftarrow} B_2,..,a \stackrel{\alpha_n}{\leftarrow} B_n\} \ \in P$ having same head $a \ $, $v_{\Im_P}(a) = (v_{\Im_P}(B_1) \wedge \alpha_1) \vee ... \vee (v_{\Im_P}(B_n) \wedge \alpha_n)$.

 \  \  \  \  \  \  \  \  \  iii.For an atom $a \in \textbf{B}_P$, and grounded rules $r_1^a: a\stackrel{\alpha_1}{\leftarrow} B_1^a, ... ,r_m^a: a\stackrel{\alpha_m}{\leftarrow} B_m^a$ and $r_1^{\neg}: \neg a\stackrel{\beta_1}{\leftarrow} B_1^{\neg}, ... ,r_n^{\neg}: \neg a \stackrel{\beta_n}{\leftarrow} B_n^{\neg}$ in $P$; 

\begin{center}

$$v_{\Im_P}(a) = max_k\{((\alpha_1 \wedge v_{\Im_P}(B_1^a))\vee..\vee(\alpha_m \wedge v_{\Im_P}(B_m^a))), \neg ((\beta_1 \wedge v_{\Im_P}(B_1^{\neg})) \vee...\vee (\beta_n \wedge v_{\Im_P}(B_n^{\neg})))\}$$; 

\end{center}

 and $v_{\Im_P}(a) \in \mathscr{T}$; where, for any $x,y \in \mathscr{T}$ $max_k\{x,y\} = x$ if $y \leq_{k_p} x$.

For the rule $a \stackrel{[1,1]}{\leftarrow} [0,1]$ and any interval $\overline c \in \mathscr{T}$, the strictly consistent interpretation $\Im = \{a: \overline c \}$ is a model for the rule. But the unique supported model is $\{a:[0,1], \neg a:[0,1]\}$.

For the rule $a \stackrel{[1,1]}{\leftarrow} [0,0]$ and any interval $\overline c \in \mathscr{T}$, the strictly consistent interpretation $\Im = \{a: \overline c \}$ is a model for the rule. But the unique supported model is $\{a:[0,0], \neg a:[1,1]\}$.

Supportedness of a model ensures that the epistemic state of the head of a rule is no more true  and no more certain than the truth and certainty degree of the epistemic state of the body combined with the rule weight. Thus while drawing inferences we do not infer truer or surer knowledge than is necessary to satisfy the rule.

Now a program may have more than one supported models.

\textbf{Proposition:} For a positive program, with no naf-literal, the unique \textit{k-minimal} supported model is the unified answer set.

\textbf{Example 1:}

Let P$_1$ be

\begin{center}

$a \ \stackrel{[1,1]}{\leftarrow} b$

$b \ \stackrel{[1,1]}{\leftarrow} a$

\end{center}

Any strictly consistent interpretation of the form $\Im = \{a:x,b:x\}; \text{such that} x \in \mathscr{T}$, is a supported model of P$_1$. However, the unique answer set is $\{a:[0,1],b:[0,1], \neg a:[0,1], \neg b:[0,1]\}$, which is the unique k-minimal supported model.

\textbf{Example 2:}

Let the following rules comprise the program P$_2$:

$r1: a \ \stackrel{[0.7,1]}{\leftarrow} b \  \  \ $ $r2: \neg a \ \stackrel{[1,1]}{\leftarrow} c \  \  \ $ $r3: a \ \stackrel{[1,1]}{\leftarrow} [0.3,0.5]$

$f1: c \ \stackrel{[1,1]}{\leftarrow} [1,1] \  \  \ $ $f2: b \ \stackrel{[1,1]}{\leftarrow} [1,1]$

This example is of particular importance to demonstrate the interaction of the epistemic states of an atom and its negative literal. During the evaluation, $a$ and $\neg a$ are considered as separate literals and their epistemic states are determined independently. Rules $r1$,$r3$ and fact $f2$ evaluate the epistemic state of atom $a$, which is $a:[0.7,1]$. Rule $r2$ and fact $f1$ evaluate the epistemic state of the literal $\neg a$, which happens to be $\neg a:[1,1]$. The second assertion for $\neg a$ is more certain (with respect to $\leq_{k_p}$) than the epistemic state for $a$. Therefore the unique answer set (i.e. the k-minimal supported model) for the program is $\{ a:[0,0], \neg a:[1,1], b:[1,1], \neg b :[0,0], c:[1,1], \neg c:[0,0]\}$.

Thus it is clear that a model may contain both the literals $a$ and $\neg a$, provided they have different degrees of certainty. The more certain valuation is added in the answer set. In other words, we would like to support decisions by taking into account the events that are have higher certainty.

\textbf{Definition 6:}
The \textbf{reduct} of a rule $r: a \stackrel{\alpha_r}{\leftarrow} b_1, b_2,...,b_m, not \ b_{m+1}, .. , not \ b_n$ with respect to an interpretation $\Im$ is $r^\Im$ and is defined as:

\begin{center}

$r^\Im : a \stackrel{\alpha_r}{\leftarrow} b_1, b_2,...,b_m, v_{\Im}(not \ b_{m+1}), .. , v_{\Im}(not \ b_n)$.

\end{center} 

i.e. $r^\Im$ doesn't contain any naf-literal in it.

The reduct of a program P $(P^\Im)$ is defined as:

\begin{center}

$P^\Im = \{ r^\Im| r \in P\}$.

\end{center}

\textbf{Definition 7:}
For an extended UnASP program $P$ an interpretation $\Im$ is said to be its \textit{answer set} if $\Im$ is the k-minimal supported model of $P^\Im$.

A UnASP program may have zero, one or more answer sets.\\

\textbf{Example 3:}

Let program P$_3$ be:

\begin{center}

$p \ \stackrel{[1,1]}{\leftarrow} not \ p$

\end{center}

The only answer set for P$_3$ is $\Im_{P_3}=\{ p:[0.5, 0.5], \neg p:[0.5, 0.5]\}$, because the reduct of $P^{\Im_{P_3}}_{3}$ becomes $p \ \stackrel{[1,1]}{\leftarrow} [0.5,0.5]$. Intuitively this means that $p$ is absolutely in the midway of the $[0,1]$ scale, i.e. $p$ is neither true nor false; which has the similar essence of its unique well-founded model assigning $p$ \textit{unknown}.

\textbf{Example 4:}

Let P$_4$ be:
 
\begin{center}
$a \ \stackrel{[1,1]}{\leftarrow} not \ b$

$b \ \stackrel{[1,1]}{\leftarrow} not \ a$

\end{center}

For any $[x,x] \in \mathscr{T}, \text{such that} \ x \in [0,1]$ the strictly consistent interpretation $\Im_{P_4} = \{a:[x,x],b:[1-x,1-x]\}$ is an answer set of P$_4$; since, $\Im_{P_4}$ is unique k-minimal supported model of the reduct $P^{\Im_{P_4}}_{4} = \{ a \ \stackrel{[1,1]}{\leftarrow} [x,x], b \ \stackrel{[1,1]}{\leftarrow} [1-x, 1-x]\}$. No interval of the form $[x,y] \text{where} \  x \neq y$ is an answer set of P$_4$. P$_4$ has infinitely many answer sets.

Fuzzy equilibrium logic \cite{schockaert2012fuzzy} produces the valuation $v(a) = [\lambda,1], v(b) = [1-\lambda,1]$, for $\lambda \in [0,1]$ as the model of this program. This solution is not in accordance to the intuitive interpretation of intervals that they are possible degrees of belief. The value 1 should not be present in both of the intervals, as both $a$ and $b$ cannot be true simultaneously.

\textbf{Example 5:}

Program P$_5$ is as follows:\\

$r1: a \ \stackrel{[1,1]}{\leftarrow} [0.9,0.9]$

$r2: \neg a \ \stackrel{[1,1]}{\leftarrow} [1,1]$\\

No strictly consistent interpretation can satisfy both the rules. Thus P$_5$ has no answer set. Intuitively the program is inconsistent since it assigns, with equal degree of certainty, intervals of very high truth degree to both $a$ and $\neg a$, which are complementary.

\textbf{Theorem 1}

Any normal UnASP program, i.e., program that does not contain any classically negated literal, is consistent and has at least one answer set.

\textbf{Proof:} Proof is presented in the Appendix section.

\subsection{Constraints:}

Some programs containing classical negation $(\neg)$ as well as negation-as-failure ($not$) give rise to unintuitive answer sets by assigning some specific values to atoms which do not occur in the head of any rule, and hence ideally should have epistemic state $[0,1]$. For instance, consider the following program:

$P = \{r1: a \longleftarrow \text{not} \ b, r2: \ \neg \ a \longleftarrow [0.6,0.6] \}$.

The only supported interpretation satisfying the program is $\{a:[0.4,0.4], b:[0.6,0.6]\}$. But this assignment of an epistemic value of high certainty to $b$ is unintuitive, since there is no rule with atom $b$ as its head, i.e., there is no source of information about $b$ is available. Intuitively the epistemic state of $b$ should be $[0,1]$ and hence the program $P$ should have no answer set. This is ensured by introducing an additional fact that acts as constraint on the possible epistemic state of $b$ in any answer set. The added fact is: 

$r3: b \longleftarrow [0,1]$ 

where rule $r3$ stops assignment of any arbitrary epistemic state to $b$. Thus this type of constraints are added to the program for every atom that does not occur in the head of any rule.

\section{Iterative approach to answer set computation:}

This section discusses an iterative approach for constructing the answer set of a restricted class of UnASP program. 

\subsection {The program transformation:}

 The first step of developing the iterative computation of the answer set of a program is to transform it in such a way that each atom appears in the head of at most one rule. For any program $P$ the corresponding transformed program is $P^*$.

\textbf{Definition 8:}

For a program $P$ and a literal $l$ the \textbf{r-join} operator for $l$ is defined as:

$r\_join(l,P) = \bigvee \{r_{body} \wedge \alpha_r | r \in P \  \text{and} \  r_{head} = l\}$

For an atom $a \in \textbf{B}_P$ the transformed rule in $P^*$ corresponding to $a$ becomes:

\begin{center}

$r^*(a) = r\_ join(a,P) \otimes_k \neg r\_ join(\neg a,P)$.

\end{center}

where, $\otimes_k$ is the \textit{knowledge aggregation operator} which takes into account the interaction of epistemic states of an atom and its corresponding negated literal based on their certainty levels. The operator $\otimes_k$ accounts for representing the nonmonotonic relation between an atom and its negation and is defined as follows:

\textbf{Definition 9:}

For two intervals $x = [x_1, x_2]$ and $y = [y_1,y_2]$ in $\mathscr{T}$; 

\ $x \otimes_k y = max_k\{x, y\}$ if $(x_2 - x_1) \neq (y_2 - y_1)$ \ \text{or} $x = y$; 

and $x \otimes_k y = [\xi, \xi]$ otherwise; where $\xi$ is a large negative number.

When for some epistemic states of $x$ and $y$,  $x \otimes_k y$ is undefined, any arbitrary large negative value $\xi$ is assigned to $x \otimes_k y$, so that occurrence of this value would denote inconsistency. 

Now, consider the transformation of a program $P$ containing the following rules:\\

$r1: a \stackrel{\alpha}{\longleftarrow} b \wedge d \  \  \ $ $r2: a \stackrel{\beta}{\longleftarrow} c \wedge e \  \  \  $ $r3: \neg a \stackrel{\gamma}{\longleftarrow} f \wedge g$

$r4: \neg a \stackrel{\delta}{\longleftarrow} k$\\

The rules are conjoined to form a single rule in the transformed program $P^*$ as follows:

\begin{center}

$r^*: a\leftarrow[((b \wedge d)\wedge \alpha) \vee ((c \wedge e) \wedge \beta)] \otimes_k \neg[((f \wedge g) \wedge \gamma) \vee (k \wedge \delta)]$.
 
\end{center}

which can be further decomposed and simplified as combination of standard DNF and CNF as follows:

\begin{center}

$r^*: a\leftarrow[(b \wedge d \wedge \alpha) \vee (c \wedge e \wedge \beta)] \otimes_k [(\neg f \vee \neg g \vee \neg \gamma) \wedge (\neg k \vee \neg \delta)]$.
 
\end{center}

Each rule, $r^*$ in the transformed program, $P^*$, combines all those rules of the original program $P$ that have either a particular atom, say $a$, or its negated literal $\neg a$ in their heads by means of $r\_join(a)$ and $r\_ join(\neg a$). One important point to note that the transformed rules in $P^*$ are \textbf{not weighted} rules any more, as a transformed rule collects weights of all associated rules in its body. 

For each atom $a$ that does not occur in the head of any rule an extra rule of the form $a \longleftarrow [0,1]$ is added to the transformed program.

The notion of interpretation for a transformed program is same as the interpretation for a UnASP program. However, since program transformation introduces an extra knowledge aggregation operator $\otimes_k$, computation of epistemic states of complex expressions using an interpretation $\Im$ is extended with the following condition (in addition to the conditions stated in section 4):

\begin{center}

$v_{\Im}(a \otimes_k b) = v_{\Im}(a) \otimes_k v_{\Im}(b)$;
\end{center}

where $a,b \in \Im^{Lit} \cup \mathscr{T}$ or $a,b$ are complex expressions constructed from grounded literals using $\wedge, \vee, \neg$ and $not$. Satisfaction of rules and model of a transformed program is defined in the same way as in Definition 4. Since there are no two rules containing same atom in their heads, condition (i) of the Definition 5 is sufficient for a model to be a supported model of the transformed program. The rest of the conditions of Definition 5 are taken care of in the construction of transformed rules. The reduct and answer set of a transformed program can be defined following Definition 6 and 7 respectively.

\textbf{Theorem 2}

An interpretation is a supported model of a UnASP program iff it is also a supported model of the transformed program corresponding to that UnASP program. 

\textbf{Proof:} Proof is presented in the Appendix section.

\textbf{Corollary:} The answer sets of a UnASP program and its transformed program are the same, provided the program is consistent.

\subsection{\textbf{Monotonic Iteration Stage (MI):}}

The first stage of computing the answer set is the monotonic iteration stage. Here an immediate consequence operator $(\Gamma)$ is used, which asserts an epistemic state to an atom $a$, if there is a rule $r^*$ in $P^*$ with $a$ in its head and the body of $r^*$ is already evaluated. 

In the transformed program each atom appears in the head of at most one rule. The iteration in the MI stage starts with the 'explicit' information given in the program in terms of \textit{facts}. As the iteration progresses, any atom, say $a$, will either remain unevaluated or will be assigned an epistemic state using the facts already known; and once an epistemic state is ascribed, it won't change throughout the iterations. That means, once a piece of information is attained it is used to deduce further information and that knowledge is never retracted or withdrawn. Hence, the reasoning process is monotonic and therefore the stage is named as Monotonic Iteration Stage.

The monotonic iteration is based on an immediate consequence operator, $\Gamma$, which is a mapping from set of partial interpretations to set of partial interpretations and can be defined as:

\textbf{Definition 10:} 
For a partial interpretation $\Im$ and a transformed program $P^*$,

$\Gamma_{P^*}(\Im) = \Im \cup \{a:\hat{c} | a \in \textbf{B}_{P^*}  \ \text{and} \ \hat{c} \in \mathscr{T} \cup [\xi, \xi]\}$

such that there is a rule $r \in P^*$ and $r_{Head} = a  \ \text{and} \  \forall l \in r_{Body}, \ l \in \Im^{Lit}$ and $\hat{c} = v_{\Im} (r_{Body}) $.

Here only strictly consistent interpretations are considered, so epistemic state of $\neg a$ need not be specified explicitly. 

The monotonic iteration takes the empty interpretation, $\Im_{MI_0} = \Phi$ as the starting point and an upward iteration is performed on the grounded transformed program $P^*$. However, as the upward iteration progresses, the program is not held constant, but it is modified or reduced in size using the information derived at each step (i.e., using those atoms that are assigned with epistemic states). Using the operator $\Gamma$, more and more information is deduced and the set of facts grows larger. The iteration proceeds as follows:

1. $\Im_{MI_0} = \Phi  \ \text{and} \ P_0 = P^*$;

2. If for some $n > 0$, there is some atom $a \in \textbf{B}_{P_n}$, such that $a:[\xi, \xi] \in \Im_{MI_n}$ then the iteration proceeds no further. Otherwise,

$\Im_{MI_{n+1}} = \Gamma_{P_{n+1}}(\Im_{MI_n})$, 

where, $P_{n+1}$ is obtained from $P_n$ by modifying it based on $\Im_{MI_n}$ as follows:

i. In $P_n$, if each literal of the body of a rule $r: a \leftarrow r_{Body}$ can be evaluated using $\Im_{MI_n}$ then effectively rule $r$ becomes of the form $r: a \leftarrow \hat{c}$, where $\hat{c}$ is an element of $\mathscr{T}$, then $\Im_{MI_{n+1}} = \Im_{MI_n} \cup \{a:v_{\Im_{MI_n}}(r_{Body})\}$ and $P_{n+1} =P_n \setminus r$.

ii. If a literal $l$ occurs in the body of a rule $r$ of $P_{n}$ and epistemic state of $l$ can be evaluated using $\Im_{MI_n}$, then $P_{n+1} = \{P_{n} \setminus r\} \cup \{r(l/\Im_{MI_n}(l)\}$, where, the notation $r(l/\hat{c})$ means the literal $l$ in $r$ is replaced be $\hat{c}$.

iii. Any rule of the form $r: a \leftarrow b_1 \wedge .. \wedge b_n \wedge [1,1] (r: a \leftarrow b_1 \wedge .. \wedge b_n \vee [0,0])$ is replaced with $r: a \leftarrow b_1 \wedge .. \wedge b_n$.

iv. Any rule of the form $r: a \leftarrow b_1 \wedge .. \wedge b_n \wedge [0,0] (r: a \leftarrow b_1 \wedge .. \wedge b_n \vee [1,1])$ is removed and $a:[0,0] (a:[1,1])$ is added to $\Im_{MI_{n+1}}$.

3. The iterations proceed until come to a fixpoint (The following theorem guarantees the existence of a fixpoint).

\textbf{Theorem 3}

For a consistent program the iterations in the MI stage terminates at a least fixpoint of $\Gamma$.

\textbf{Proof:} Proof is presented in the Appendix section.

The final interpretation $\Im_{MI_{\infty}}$ and the corresponding final program $P_{MI}$ are the output of the Monotonic Iteration stage and they are subjected to further modifications in next stage.

\textbf{Example 6:} Let P be a grounded logic program (the propositional atoms are actually short-hand representation of grounded predicate atoms):\\

\{$r1: p \stackrel{[0.7,1]}{\longleftarrow} q, \text{not} \  s$, \  \ $r2: p \stackrel{[0.3,0.3]}{\longleftarrow} r, \text{not}  \ t$, \  \ $r3: \neg p \stackrel{[1,1]}{\longleftarrow} s$,

$r4: \neg p \stackrel{[1,1]}{\longleftarrow} t$, \  \  \  $r5: m \stackrel{[0.6,0.8]}{\longleftarrow} n$, \  \  \ $r6: s \stackrel{[1,1]}{\longleftarrow} m$,

$r7: a \stackrel{[1,1]}{\longleftarrow} b, p$, \  \  \  $r8: b \stackrel{[1,1]}{\longleftarrow} \text{not} \ a$, \  \  \ $r9: b \stackrel{[1,1]}{\longleftarrow} g$,

$r10: d \stackrel{[1,1]}{\longleftarrow} \neg g, a$, \  \  \ $r11: e \stackrel{[1,1]}{\longleftarrow} d, w$, \  \  \ $r12: f \stackrel{[1,1]}{\longleftarrow} \text{not} \ e$,

$r13: g \stackrel{[1,1]}{\longleftarrow} \neg c, f$, \  \  \ $r14: c \stackrel{[1,1]}{\longleftarrow} h$, \  \  \ $r15: h \stackrel{[0.7,1]}{\longleftarrow} k$,

$r16: i \stackrel{[1,1]}{\longleftarrow} \text{not} \ h$, \  \  \ $r17: k \stackrel{[1,1]}{\longleftarrow} \neg j$, \  \  \ $r18: j \stackrel{[0.8, 0.8]}{\longleftarrow} i, \text{not} \ s$,

$r19: \neg j \stackrel{[1,1]}{\longleftarrow} s$, \  \  \ $r20: x \stackrel{[0.2, 0.3]}{\longleftarrow} v$, \  \  \ $r21: v \stackrel{[1,1]}{\longleftarrow} x$,

$r22: v \stackrel{[1,1]}{\longleftarrow} \neg u$, \  \  \ $r23: w \stackrel{[1,1]}{\longleftarrow} x$, \  \  \ $r24: u \stackrel{[0.5, 0.8]}{\longleftarrow} w$,

$r25: y \stackrel{[1,1]}{\longleftarrow} \text{not} \ z$, \  \  \ $r26: z \stackrel{[1,1]}{\longleftarrow} \text{not} \ y$, \  \  \ $r27: l \stackrel{[0.4, 0.6]}{\longleftarrow} z$,

$f1: q  \stackrel{[1,1]}{\longleftarrow} [0.7, 0.7]$, \  \  \ $f2: r \stackrel{[1,1]}{\longleftarrow} [0.5, 0.5]$, \  \  \ $f3: n \stackrel{[1,1]}{\longleftarrow} [0.7, 0.9]\}$.

The corresponding transformed program is $P^*$ as follows:\\

\{$r1^*:  p \longleftarrow ((q \wedge \text{not} \ s \wedge [0.7,1]) \vee (r \wedge \text{not} \ t \wedge [0.3,0.3])) \otimes_k (\neg s \wedge \neg t) $,

$r2^*: m \longleftarrow n \wedge [0.6, 0.8]$, \  \  \ $r3^*: s \longleftarrow m$, \  \ $r4^*: a \longleftarrow b \wedge p$, 

$r5^*: b \longleftarrow \text{not} \ a \vee g$, \  \ $r6^*: d \longleftarrow \neg g \wedge a$, \  \  \ $r7^*: e \longleftarrow d \wedge w$, 

$r8^*: f \longleftarrow \text{not} \ e$, \  \  \ $r9^*: g \longleftarrow \neg c \wedge f$, \  \  \ $r10^*: c \longleftarrow h$, 

$r11^*: h \longleftarrow k \wedge [0.7,1]$, \  \  \ $r12^*: i \longleftarrow \text{not} \ h$, \  \  \ $r13^*: k \longleftarrow \neg j$,

$r14^*: j \longleftarrow ([0.8,0.8] \wedge i \wedge \text{not} \ s) \otimes_k \neg s$,

$r15^*: x \longleftarrow  v \wedge [0.2, 0.3]$, \  \  \ $r16^*: v \longleftarrow x \vee \neg u$, \  \  \ $r17^*: w \longleftarrow x$,

$r18^*: u \longleftarrow w \wedge [0.5,0.8]$, \  \ $r19^*: y \longleftarrow \text{not} \ z$, \  \ $r20^*: z \longleftarrow \text{not} \ y$,

$r21^*: l \longleftarrow z \wedge [0.4, 0.6]$, 

$f1^*: q  \longleftarrow [0.7, 0.7]$, \  \ $f2^*: r \longleftarrow [0.5, 0.5]$, \  \ $f3^*: n \longleftarrow [0.7, 0.9]$,

$f4^*: t \longleftarrow [0,1]\}$.

\vspace{5mm}

The monotonic Iterations progress on the transformed program $P^*$ as follows:

1. $\Im_{MI_0} = \Phi, P_0 = P^*$\\

2. $\Im_{MI_1} = \Gamma (\Im_{MI_0})$

    \  \  \ $= \{ q:[0.7,0.7], r:[0.5,0.5], n:[0.7,0.9], t:[0,1]\}$.

	 \  \  \ $P_1 = P^*$.\\

3. $\Im_{MI_2} = \Gamma (\Im_{MI_1})$.

   \  \  \ $= \Im_{MI_1} \cup \{ m:[0.42,0.56]\}$.

   \  \  \ $P_2 = P_1 \setminus \{r1^*, r2^*, r3^*\} \cup \{r1^*(q/[0.7,0.7],r/[0.5,0.5],t/[0,1]), r3^*(m/[0.42,0.56])\}.$
	
where, 

$r1^*(q/[0.7,0.7],r/[0.5,0.5],t/[0,1]): p \longleftarrow (([0.49,0.7] \wedge \text{not} \ s) \vee [0.15,0.15]) \otimes_k \neg(s \vee [0,1])$,

$r3^*(m/[0.42,0.56]): s \longleftarrow [0.42,0.56]$.\\

4. $\Im_{MI_3} = \Gamma (\Im_{MI_2})$

   \  \  \ $ = \Im_{MI_2} \cup \{s:[0.42,0.56]\}$.

   \  \  \ $P_3 = P_2 \setminus \{r1^*, r3^*, r14^*\} \cup \{r1^*(s/[0.42,0.56]), r14^*(s/[0.42,0.56])\}$.
	
	where, 
	
	$r1^*(s/[0.42,0.56]):p \longleftarrow (([0.49,0.7] \wedge [0.58,0.58]) \vee [0.15,0.15]) \otimes_k \neg([0.42,0.56] \vee [0,1])$,
	
	$r14^*(s/[0.42,0.56]):j \longleftarrow ([0.8,0.8] \wedge i \wedge [0.58,0.58]) \otimes_k [0.44,0.58]$.\\

5. $\Im_{MI_4} = \Gamma (\Im_{MI_3})$ 

   \  \  \ $= \Im_{MI_3} \cup \{p:[0.3916,0.495]\}$.

   \  \  \ $P_4 = P_3 \setminus \{r1^*, r4^*\} \cup \{r4^*(p/[0.3916,0.495])\}$.
	
where,

  $r4^*(p/[0.928,0.941]): a \longleftarrow b \wedge [0.3916,0.495]$.
	
6. $\Im_{MI_5} = \Gamma (\Im_{MI_4}) = \Im_{MI_4}$\\

	The MI stage stops here, because further iterations doesn't modify the program, nor the deduced set of facts is modified.
	
	$\Im_{MI_{\infty}} = \Im_{MI_4}$
	
	$ \  \  \ =\{q:[0.7,0.7], r:[1,1], n:[0.7,0.9], t:[0,1], m:[0.42,0.56], s:[0.42,0.56], p:[0.3916,0.495]\}$.\\
	
	$P_{MI} = P_4 =$
	
	\{$r1_{MI}: a \longleftarrow b \wedge [0.3916,0.495]$,$ \  \  \  \   \  \ \  \  \  \  r2_{MI}: b \longleftarrow \text{not} \ a \vee g$,

$r3_{MI}: d \longleftarrow \neg g \wedge a$, $ \  \  \  \  \  \  \   \ \  \  \  \  \  \  \  \  \  \  \  \  \  \  \ r4_{MI}: e \longleftarrow d \wedge w$,

$r5_{MI}: f \longleftarrow \text{not} \ e$, $ \  \  \  \  \  \  \  \  \  \  \  \  \  \  \  \  \  \  \  \  \  \  \  \ r6_{MI}: g \longleftarrow \neg c \wedge f$,

$r7_{MI}: c \longleftarrow h$, $ \  \  \  \  \  \  \  \  \  \  \  \  \  \  \  \  \  \  \  \  \  \  \  \  \  \  \  \  \ r8_{MI}: h \longleftarrow k \wedge [0.7,1]$,

$r9_{MI}: i \longleftarrow \text{not} \ h$, $ \  \  \  \  \  \  \  \  \  \  \  \  \  \  \  \  \  \  \  \  \  \  \  \  \  \ r10_{MI}: k \longleftarrow \neg j$,

$r11_{MI}: j \longleftarrow ([0.464,0.464] \wedge i) \otimes_k [0.44,0.58]$, $ \  \  \  \  \  \ r12_{MI}: x \longleftarrow v \wedge [0.2, 0.3]$,

$r13_{MI}: v \longleftarrow x \vee \neg u$, $ \  \  \  \  \  \  \  \  \  \  \  \  \  \  \  \  \  \  \  \  \  \  \  \  \  \  \  \  \  \  \  \  \  \ r14_{MI}: w \longleftarrow x$,

$r15_{MI}: u \longleftarrow w \wedge [0.5,0.8]$, $ \  \  \  \  \  \  \  \  \  \  \  \  \  \  \  \  \  \  \  \  \  \  \  \  \  \  \  r16_{MI}: y \longleftarrow \text{not} \ z$,

$r17_{MI}: z \longleftarrow \text{not} \ y$, $ \  \  \  \  \  \  \  \  \  \  \  \  \  \  \  \  \  \  \  \  \  \  \  \  \  \  \  \  \  \  \  \  \  \ r18_{MI}: l \longleftarrow z \wedge [0.4, 0.6]$\}.\\

The output of the Monotonic Iteration is a reduced program, where, the rules that are already evaluated, are absent. This reduced program $P_{MI}$ consists of rules where the epistemic state of any atom is in some way self-dependent or dependent on epistemic states of some self-assessing atoms. So to evaluate the epistemic states of atoms in the head of such rules 'Guess and Check' approach is taken. An initial guess for the epistemic states is made and then those epistemic states are put in the bodies of rules and the heads are evaluated. If the evaluated epistemic states match with the guessed epistemic states then the guess was a stable valuation. This stable valuation is to be obtained by using iterations starting from an initial guess, $[0,1]$, which signifies that nothing is known about the corresponding atom. This iteration continues until the evaluated epistemic state is "sufficiently close" to the guessed interval. 

Next subsections describe the iteration method, its analysis and sufficient condition for the convergence of iterations to a fixed-point. 

\subsection{Programs represented by Dependency Graph and Program Splitting:}

To extract information about the interrelation of rules in $P_{MI}$, detection of self-assertive set of rules (i.e. cycles) and their connections, a Dependency Graph (DG) is constructed from $P_{MI}$. It is a directed graph with weighted edges. 

\textbf{Definition 11:}

A dependency Graph(DG) for a transformed program $P$ is defined as a triplet,

			$ \  \  \  \  \  \  \  \  \ DG = <V, E, W>$
			
			where, 
			
			$V$ is set of vertices; $V = \textbf{B}_{P} \cup \mathscr{T} \cup \{\wedge, \vee, \otimes_k\}$\\
			
			$E = \{<p,q> | p,q \in V\}$.
			An edge can be between two literals or between a literal and a logical connective (i.e. $\vee, \wedge, \otimes_k)$ depending on the rules in the program $P$.\\
						
			$W$ is a \textit{partial mapping} from the set of edges $E$ to the set $\{-1,\neg\}$ such that,
						
						 $ \  \  \  \  \  \  W(e) = -1$ if $e \in E \  \text{and} \ e = <p,q>$ such that $e$ connects a literal $not \ p$ to another vertex $q$;
						
						 $ \  \  \  \  \  \  W(e) = \neg $ if $e \in E \  \text{and} \ e = <p,q>$ such that $e$ connects a literal $\neg \ p$ to another vertex $q$;
						
						 $ \  \  \  \  \  \  W(e)$ is unspecified otherwise.

\begin{figure}
\centering
\begin {tikzpicture}[-latex ,auto ,node distance =1.5 cm and 2 cm ,on grid ,
semithick ,
state/.style ={ circle,
draw, text=black , minimum width = 0.5 cm}]

\node[state] (C){$c$};
\node[state] (H)[below =of C]{$h$};
\node[state] (CH)[below right=of H]{$\wedge$};
\node (nh)[left =of CH]{$[0.7,1]$};
\node[state] (I) [above right =of H]{$i$};
\node[state] (CJ) [right =of I]{$\wedge$};
\node (ncj) [below =of CJ]{$[.464,.464]$};
\node[state] (KJ) [below right =of CJ]{$\otimes_k$};
\node (nj) [right =of KJ]{$[.44,.58]$};
\node[state] (J) [below =of KJ]{$j$};
\node[state] (K) [left =of J]{$k$};

\node[state] (CG) [above =of C]{$\wedge$};
\node[state] (G) [right =of CG]{$g$};
\node[state] (DB) [right =of G]{$\vee$};
\node[state] (B)[right =of DB]{$b$};
\node[state] (CA) [above =of B]{$\wedge$};
\node (na) [right =of CA]{$[0.3916,0.495]$};
\node[state] (A) [left =of CA]{$a$};
\node[state] (CD) [left =of A]{$\wedge$};
\node[state] (D) [left =of CD]{$d$};
\node[state] (CE) [left =of D]{$\wedge$};
\node[state] (E) [left =of CE]{$e$};
\node[state] (F) [below =of E]{$f$};

\node[state] (W) [above right =of CE]{$w$};
\node[state] (X) [left =of W]{$x$};
\node[state] (CX) [left =of X]{$\wedge$};
\node[state] (CU) [right =of W]{$\wedge$};
\node (nu) [right =of CU]{$[.5,.8]$};
\node[state] (U) [above =of CU]{$u$};
\node[state] (DV) [left =of U]{$\vee$};
\node[state] (V) [left =of DV]{$v$};

\node (nx) [above =of CX]{$[.2,.3]$};

\node[state] (L) [right =of U]{$l$};
\node[state] (CL) [right =of L]{$\wedge$};
\node (nl) [below =of CL]{$[.4,.6]$};
\node[state] (Z) [right =of CL]{$z$};
\node[state] (Y) [right =of Z]{$y$};

\path (H) edge (C);
\path (nh) edge (CH);
\path (CH) edge (H);
\path (H) edge [bend left =25] node[above =.10cm]{$-1$} (I);
\path (I) edge (CJ);
\path (ncj) edge (CJ);
\path (CJ) edge (KJ);
\path (nj) edge (KJ);
\path (KJ) edge (J);
\path (J) edge node[below =0.10cm] {$\neg$} (K);
\path (K) edge (CH);

\path (C) edge node[right =.10 cm]{$\neg$} (CG);

\path (W) edge (CE);
\path (CE) edge (E);
\path (E) edge node[left =.10 cm]{$\neg$} (F);
\path (F) edge (CG);
\path (CG) edge (G);
\path (G) edge (DB);
\path (DB) edge (B);
\path (B) edge [bend right =25] (CA);
\path (CA) edge (A);
\path (A) edge (CD);
\path (A) edge node[left =.10 cm]{$-1$} (DB);
\path (G) edge node[right =.10 cm]{$\neg$} (CD);
\path (CD) edge (D);
\path (D) edge (CE);
\path (na) edge (CA);

\path (W) edge (CU);
\path (CU) edge (U);
\path (X) edge (W);
\path (X) edge [bend right =25](DV);
\path (V) edge [bend right =25] (CX);
\path (CX) edge (X);
\path (U) edge node[above =0.10cm] {$\neg$} (DV);
\path (DV) edge (V);
\path (nu) edge (CU);
\path (nx) edge (CX);

\path (CL) edge (L);
\path (nl) edge (CL);
\path (Z) edge (CL);
\path (Z) edge [bend right =25] node[below =.10 cm]{$-1$} (Y);
\path (Y) edge [bend right =25] node[above =.10 cm]{$-1$} (Z);
\end{tikzpicture}
\caption{Dependency graph of the program $P_{MI}$}
\label{figure:THE example}
\end{figure}

The dependency graph of the program $P_{MI}$ is shown in figure \ref{figure:THE example}. This graph is referred to as $G$ for further references.

On the dependency graph $G$, Kosaraju's algorithm is applied to detect the \textbf{Strongly-Connected Components} in it. Then the Component Graph ($G^{SCC}$) is constructed by merging all strongly connected nodes into a single node. This $G^{SCC}$ is a Directed Acyclic Graph (DAG). The nodes in $G^{SCC}$ are sorted using a Topological Sort, which actually determines the inter-dependencies of rules in the program $P_{MI}$ and determines a sequence in which the calculation of epistemic states of atoms would proceed.

 \begin{figure}
\centering
\begin {tikzpicture}[-latex ,auto ,node distance =2 cm and 2cm ,on grid ,
semithick ,
state/.style ={ circle,
draw, text=black , minimum width = 0.5 cm}]

\node[state] (X){$uvxw$};
\node (nx) [left =of X]{$[.2,.3]$};
\node (nu) [right =of X]{$[.5,.8]$};

\node (L) [right =of nu]{$l$};
\node[state] (CL) [right =of L]{$\wedge$};
\node (nl) [below =of CL]{$[.4,.6]$};
\node[state] (Z) [right =of CL]{$yz$};

\node[state] (G) [below =of X]{$gbadef$};
\node (na) [right =of G]{$[.392,.495]$};

\node[state] (C) [below =of G]{$c$};

\node[state] (H) [below =of C]{$hijk$};
\node (nj1) [left =of H]{$[.464,.464]$};
\node (nj2) [right =of H]{$[.44,.58]$};
\node (nk) [below =of H]{$[0.7,1]$};

\path (H) edge (C);
\path (nj1) edge (H);
\path (nj2) edge (H);
\path (nk) edge (H);

\path (C) edge (G);

\path (X) edge (G);
\path (na) edge (G);

\path (CL) edge (L);
\path (nl) edge (CL);
\path (Z) edge (CL);

\path (nx) edge (X);
\path (nu) edge (X); 

\end{tikzpicture}
\caption{Component graph of $G^{SCC}$}
\label{figure:GSCC}
\end{figure}
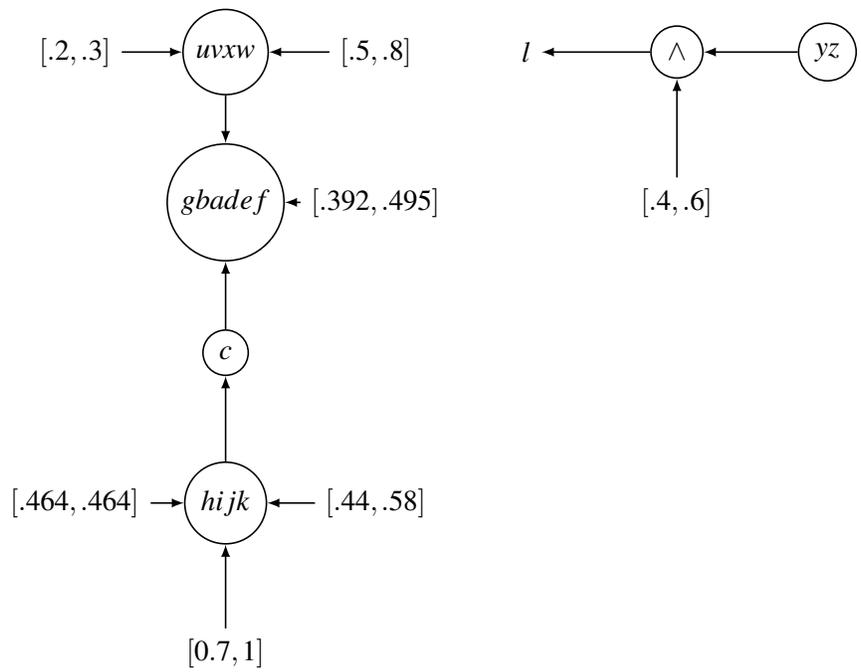

From DG $G$ the component graph $G^{SCC}$is shown in figure \ref{figure:GSCC}. One of the many possible topological sorts of $G^{SCC}$ is 

\begin{center}

$hijk, uvxw, c, gbadef, yz, l$

\end{center}

where, constant nodes (e.g. $[.44,.58]$) and nodes for logical operators(e.g. $\wedge$) are implicit. Each node in the topologically sorted list is either a singleton, i.e., a constant or connective, or an SCC with multiple nodes. As can be seen from figure \ref{figure:THE example} that each non-singleton SCC may be composed of a simple cycle or may contain many interconnected simple cycles.

\subsection{Nonmonotonic Evaluation:}

In the Monotonic Iteration stage epistemic states of some of the atoms (and literals) are evaluated using the given facts. The output program of the MI stage, $P_{MI}$, contains rules with cyclic dependencies. To evaluate the epistemic states of atoms in the head of such rules 'Guess and Check' approach is taken. An initial guess for the epistemic states is made and then those epistemic states are put in the bodies of rules and the heads are re-evaluated. If the evaluated epistemic states match with the guessed epistemic states then the guess was a stable valuation. The initial epistemic states of the atoms are selected depending on the nature of the program that is being evaluated. Based on the program under consideration and the initial values the nonmontonic evaluation can be of two types: 1. Nonmonotonic Iteration (NMI) and 2. Branch-and-bound.

In case of NMI iteration over some program segment, the interval $[0,1]$ is chosen as the initial epistemic state of the atoms to be evaluated. The stable valuation is to be obtained by iterations starting from the initial guess, $[0,1]$, which signifies that nothing is known about the corresponding atom. This iteration continues until the evaluated epistemic state is "sufficiently close" to the guessed interval. NMI iteration is applicable for program segments having a unique answer set. 

For program segments, having multiple answer sets, NMI can't produce all the answer sets. In that case Branch-and-bound is applied. Here, atoms are initialized with intervals of the form $[x,x]$, s.t. $x \in [0,1]$. Then this value is iterated over the program segment for a single time and the evaluated epistemic state is compared with the guessed value for convergence.

\subsubsection{Assumption set of a program:}

	A program segment represented by a particular SCC in the dependency graph is denoted by $P_{scc}$. Instead of initializing all the atoms in $P_{scc}$ and comparing their evaluated values for checking convergence, some 'special' atoms are chosen from the set of atoms in $B_{P_{SCC}}$ and the iterations are observed from the perspective of these 'special' atoms \textit{only}. Once stable valuation for these atoms are attained, the stable epistemic states of the rest of the atoms in $B_{P_{SCC}}$ can be calculated. The set of these chosen 'special' atoms is called the \textbf{Assumption Set}. 

\textbf{Definition 12:}

An Assumption Set ($As_P$) for a set of rules or program  $P$, whose dependency graph is a strongly connected component, is the set of atoms such that, the epistemic states of all other atoms present in the set of rules can be uniquely determined if the epistemic states of the elements of the assumption set are specified. Each element in the assumption set is called a \textit{chosen element} or a \textit{chosen atom}.

The assumption set is the set of atoms using which the interconnected cycles of an SCC can be 'unfurled' into linear forward paths, along which the computation progresses and epistemic states of the chosen elements propagate through the SCC and other atoms are evaluated. This unfurled graph obtained from the SCC is called the \textbf{value-propagation graph} (vpg). This vpg provides a way to mathematically analyse the nature of non-monotonic evaluations (especially of NMI iterations) and investigate the terminating conditions. The assumption set has to be chosen in such a way so that the resulting vpg is cycle-free;  wrong choice may lead to backward paths in the vpg, which is unwanted.

\textbf{Finding assumption set of a program from its dependency graph:}

Given a program segment which corresponds to a non-singleton SCC in the dependency graph, Johnson's Algorithm is applied to detect simple cycles in that SCC. The set of simple cycles in the SCC as obtained by Johnson's algorithm is denoted by $J_{scc}$. With the set of atoms in the SCC, i.e. $\textbf{B}_{P_{SCC}}$, and the set of simple cycles, $J_{scc}$, an \textit{intersection table} is constructed with elements of $\textbf{B}_{P_{SCC}}$ as column heads and elements of $J_{scc}$ as row heads and putting a tick ($\checkmark$) in the cell $(C,a)$, if the atom $a$ occurs in the cycle $C$. If an atom, occurring in the $i^{th}$ column is chosen then all the cycles that have a $\checkmark $ in the $i^{th}$ column will be unfurled in the vpg. So, in order to construct the assumption set the set of atoms is picked up based on fulfillment of the following criteria:

1. the set of chosen atoms has to be such that \textit{all} the cycles from  $J_{scc}$ are unfurled in vpg.

2. for each chosen element, there has to be a cyclic path in the SCC starting and ending with the chosen atom and that does not include any other chosen element.

3. the assumption set has to be as small as possible.

Some additional criteria may have to be imposed depending on whether NMI or branch-and-bound is chosen.

Condition 2, mentioned above, ensures that for each chosen element, say $a$, there is a path from $a_{n-1}$ to $a_n$ in the vpg that is obtained by unfurling all the cycles from the SCC that start and end at $a$ and doesn't have any other chosen element on it. This path shows how value of $a$ at $n^{th}$ iteration is computed from $(n-1)^{th}$ iteration. Such a path is called \textbf{value-propagation path}(vpp) of $a$. 

Clearly, the assumption set for any program may not be unique.

The dependency graph and concept of studying iterations through elements of the assumption set offer a new perspective to study the nature and convergence conditions of iterations.\\

\subsubsection{NMI Iteration:}

Based on the concept of assumption set for a set of rules or program segment $P$, the steps of NMI iterations are as follows:

1. The iteration starts with an interpretation $\Im_{NMI}^0 = \{a:[0,1]| \forall a \in As_P\}$.\\

2. Having $\Im_{NMI}^n$, $\Im_{NMI}^{n+1}$ is obtained in two steps.

 \  \  \  \  \  \  \ i. $P$ is modified to construct $P(\Im_{NMI}^n)$, such that for any rule $r \in P$ and for any $a \in As_P$ if $a \in r_{Body}$ it is replaced with $\Im_{NMI}^n(a)$.\\

\  \  \  \  \  \  ii. Monotonic iterations are performed on $P(\Im_{NMI}^n)$ using the immediate consequence operator $\Gamma$ (Definition 10); with

$ \  \  \  \  \  \  \  \  \  \  \  \  \ \Im_0 = \Phi$;

$ \  \  \  \  \  \  \  \  \  \  \  \  \ \Im_m = \Gamma_{P(\Im_{NMI}^n)}(\Im_{m-1})$, for some $m \in \mathbb{N}$

until a fixpoint $\Im_{\infty}$ is obtained.\\

$ \  \  \  \  \  \  \  \  \  \  \  \  \ \Im_{NMI}^{n+1} = \{a:\Im_{\infty}(a) | a \in As_P\}$.\\

3. For each element $a \in AS_P$, whose epistemic state at n$^{th}$ iteration is denoted by $[a_{1_n},a_{2_n}]$, $|\Im_{NMI}^{n+1}(a) - \Im_{NMI}^n(a)|$ can be represented by a vector $D_a = [|a_{1_{n+1}} - a_{1_n}| \  \ |a_{2_{n+1}} - a_{2_n}|]$. Step 2 is repeated until $\forall a \in AS_P; ||D_a||_\infty < \epsilon$ for some pre-decided $\epsilon > 0$, i.e. difference in magnitude of both of the upper bound and the lower bound is less than $\epsilon$.\\

\begin{figure}[ht]
    \centering
\begin{tabular}{c@{\qquad}c}
    \begin {tikzpicture}[-latex ,auto ,node distance =1 cm and 1.5 cm ,on grid ,
semithick ,
state/.style ={ circle,
draw, text=black , minimum width = 0.5 cm}]
\node[state] (C){$c$};
\node[state] (CA)[right =of C]{$\wedge$};
\node[state] (A)[below =of CA]{$a$};
\node (nA)[right =of CA]{$[0.6,0.8]$};
\node[state] (DB) [below left =of C]{$\vee$};
\node[state] (B)[above =of DB]{$b$};
\node[state] (CD) [below =of A]{$\wedge$};
\node[state] (G) [below =of DB]{$g$};
\node[state] (CG) [below =of G]{$\wedge$};
\node (ng) [left =of CG]{$[0.3,0.7]$};
\node[state] (D) [below =of CD]{$d$};
\node[state] (CE) [below =of D]{$\wedge$};
\node (ne) [right =of CE]{$[0.9,1]$};
\node[state] (E) [left =of CE]{$e$};
\node[state] (F) [left =of E]{$f$};

\path (C) edge (CA);
\path (CA) edge (A);
\path (nA) edge (CA);
\path (A) edge node[above =0.07 cm] {$-1$}(DB);
\path (DB) edge (B);
\path (B) edge (C);
\path (G) edge (DB);
\path (A) edge (CD);
\path (G) edge node[above =0.07 cm] {$\neg$} (CD);
\path (CD) edge (D);
\path (D) edge (CE);
\path (CE) edge (E);
\path (E) edge node[above =0.07 cm] {$\neg$} (F);
\path (F) edge (CG);
\path (CG) edge (G);
\path (ng) edge (CG);
\path (ne) edge (CE);

\end{tikzpicture} 
    &
    \begin{tabular}{l | c c c c c c c c}
    \toprule
Simple Cycles  & $a$ & $b$ & $c$ & $d$ & $e$ & $f$ & $g$\\ \midrule
 $abca$              & $\checkmark$     & $\checkmark$     & $\checkmark$ &  &  &  & \\
 $adefgbca$              & $\checkmark$     & $\checkmark$     & $\checkmark$ & $\checkmark$ & $\checkmark$ & $\checkmark$ & $\checkmark$\\
 $gdefg$               &      &      &    & $\checkmark$ & $\checkmark$ & $\checkmark$ & $\checkmark$\\
 \bottomrule
    \end{tabular}
\end{tabular}
\caption{Dependency graph of $P_{Ex.7}$ and the Intersection table}
    \label{figure:example2}
\end{figure}
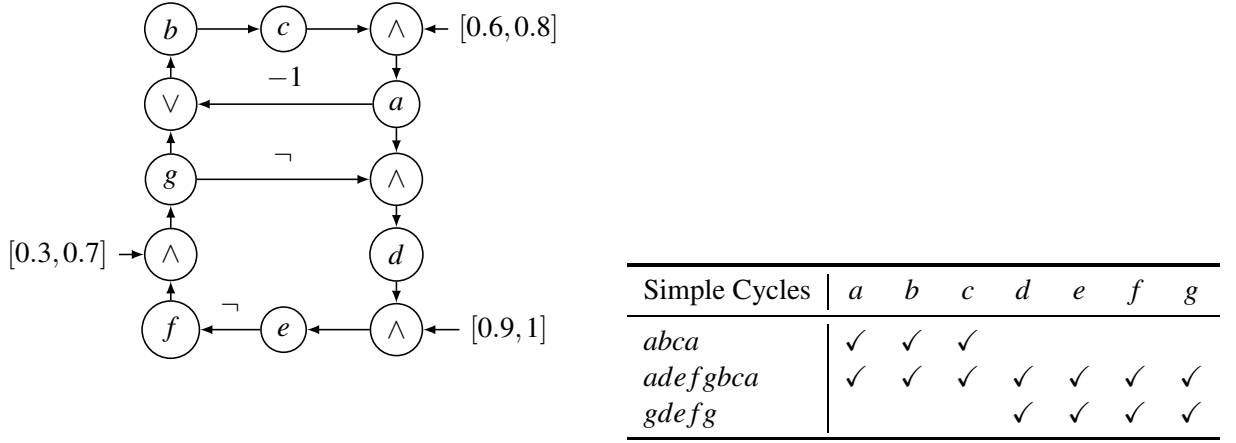

\textbf{Example 7:} Lets consider the program segment, whose dependency graph and the corresponding intersection table are shown in Figure \ref{figure:example2}. The corresponding set of rules are:

$P_{Ex.7} = \{a \leftarrow c \wedge [0.6,0.8];$

$ \  \  \ \  \  \  \  \  c \leftarrow b;$

$ \  \  \ \  \  \  \   \ b \leftarrow not \ a \vee g;$

$ \  \  \ \  \  \  \  \  d \leftarrow \neg g \wedge a;$

$ \  \  \ \  \  \  \  \  e \leftarrow d \wedge [0.9,1];$

$ \  \  \ \  \  \  \  \  f \leftarrow \neg e;$

$ \  \  \ \  \  \  \  \  g \leftarrow f \wedge [0.3,0.7]\}.$

\begin{figure}
\centering
\begin {tikzpicture}[-latex ,auto ,node distance =1.5 cm and 1.5 cm ,on grid ,
semithick ,
state/.style ={ circle,
draw, text=black , minimum width = 0.5 cm}]

\node[state] (A){$a_{n-1}$};
\node[state] (DB)[above right=of A]{$\vee$};
\node[state] (B) [right =of DB]{$b_n$};
\node[state] (C)[right=of B]{$c_n$};
\node[state] (CAo) [right =of C]{$\wedge$};
\node[state] (Ao)[below right=of CAo]{$a_n$};
\node (nA)[below =of CAo]{$[0.6,0.8]$};

\node[state] (G)[below=of A]{$g_{n-1}$};
\node[state] (CD)[below right=of G]{$\wedge$};
\node[state] (D)[right=of CD]{$d_n$};
\node[state] (CE)[right=of D]{$\wedge$};
\node[state] (E)[right=of CE]{$e_n$};
\node (nE)[below =of CE]{$[0.9,1]$};
\node[state] (F)[right=of E]{$f_n,-$};
\node[state] (CGo)[above right =of F]{$\wedge$};
\node (nGo)[left=of CGo]{$[0.3,0.7]$};
\node[state] (Go)[above=of CGo]{$g_n$};

\path (A) edge  [bend left =25] node[above =0.15 cm] {$-1$} (DB);
\path (DB) edge (B);
\path (B) edge (C);
\path (C) edge (CAo);
\path (CAo) edge [bend left =25] (Ao);
\path (nA) edge (CAo);

\path (A) edge (CD);
\path (G) edge (DB);

\path (G) edge [bend right =25] node[below =0.15 cm]{$\neg$} (CD);
\path (CD) edge (D);
\path (D) edge (CE);
\path (CE) edge (E);
\path (nE) edge (CE);
\path (E) edge node[below =0.15 cm]{$\neg$} (F);
\path (F) edge (CGo);
\path (CGo) edge (Go);
\path (nGo) edge (CGo);

\end{tikzpicture}
\caption{Value Propagation Graph for $P_{Ex.7}$}
\label{figure:second example propagation}
\end{figure}
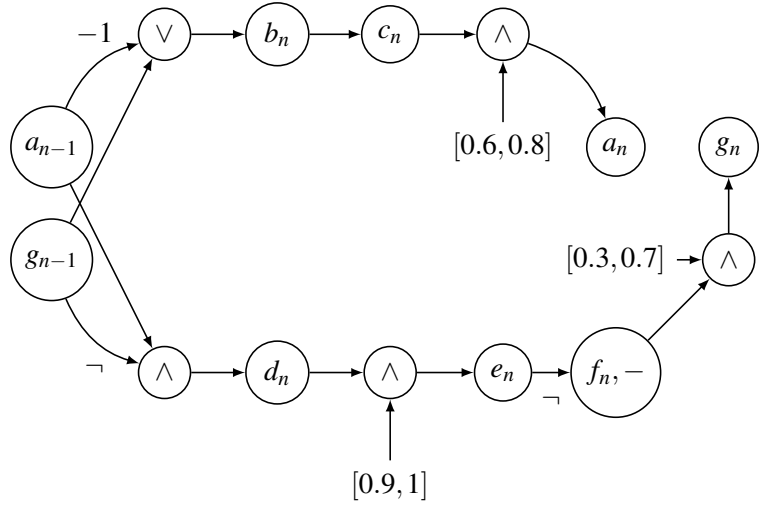

Say, the chosen assumption set is $As_{P_{Ex.7}} = \{a,g\}$. It is clear from the intersection table that these two chosen elements cover all the cycles present, as can be seen in the vpg in Figure \ref{figure:second example propagation}. The vpg shows how the epistemic states of the chosen elements at the $n^{th}$ iteration can be obtained from the epistemic states of the $(n-1)^{th}$ iteration. Say $\epsilon$ is chosen to be 0.009\\
 
1. $\Im_{NMI}^0 = \{a:[0,1], g:[0,1]\}$;

2. $P_{Ex.8}(\Im_{NMI}^0) = \{a \  \leftarrow \ c \wedge [0.6,0.8];$

$ \  \   \  \  \  \  \  \  \  \  \  \  \  \ c \  \leftarrow \ b;$

$ \  \  \  \  \  \   \  \  \  \  \  \  \  \ b \  \leftarrow \ \text{not} \ [0,1] \vee [0,1];$

$ \  \   \  \  \  \  \  \  \  \  \  \  \  \ d \  \leftarrow \ \neg [0,1] \wedge [0,1];$

$ \  \  \  \  \  \  \  \  \  \  \  \  \  \  e \  \leftarrow \ d \wedge [0.9,1];$

$ \  \  \  \  \  \  \  \  \  \  \  \  \  \  f \  \leftarrow \ \neg e;$

$ \  \  \  \  \  \  \  \  \  \  \  \  \  \  g \  \leftarrow \ f \wedge [0.3,0.7];\}$\\

The monotonic iteration using $\Gamma$ goes as follows:\\

$\Im_{01} = \Gamma_{P_{Ex.7}(\Im_{NMI}^0)}(\Phi) = \{b:[1,1], d:[0,1]\}$;

$\Im_{02} = \Gamma_{P_{Ex.7}(\Im_{NMI}^0)}(\Im_{01}) = \Im_{01} \cup \{c:[1,1], e:[0,1]\}$;

$\Im_{03} = \Gamma_{P_{Ex.7}(\Im_{NMI}^0)}(\Im_{02}) = \Im_{02} \cup \{a:[0.6,0.8], f:[0,1]\}$;

$\Im_{04} = \Gamma_{P_{Ex.7}(\Im_{NMI}^0)}(\Im_{03}) = \Im_{03} \cup \{g:[0,0.7]\}$;

$\Im_{0\infty} = \Im_{04}$.\\

$\Im_{NMI}^1 = \{a:[0.6,0.8], g:[0,0.7]\}$.\\

Since $||D_a||_\infty = max\{0.6,0.2\} = 0.6 > \epsilon$ and $||D_g||_\infty = max\{0,0.3\} = 0.3 > \epsilon$; the iteration is continued.

3. The immediate consequence operator $\Gamma$ iterated on $P_{Ex.7}(\Im_{NMI}^1)$ proceeds as follows:\\

$\Im_{11} = \Gamma_{P_{Ex.7}(\Im_{NMI}^1)}(\Phi) = \{b:[0.4, 0.82], d:[0.18,0.8]\}$;

$\Im_{12} = \Gamma_{P_{Ex.7}(\Im_{NMI}^1)}(\Im_{11}) = \Im_{11} \cup \{c:[0.4, 0.82], e:[0.162,0.8]\}$;

$\Im_{13} = \Gamma_{P_{Ex.7}(\Im_{NMI}^1)}(\Im_{12}) = \Im_{12} \cup \{a:[0.24,0.656], f:[0.2,0.838]\}$;

$\Im_{14} = \Gamma_{P_{Ex.7}(\Im_{NMI}^0)}(\Im_{13}) = \Im_{13} \cup \{g:[0.06,0.5866]\}$;

$\Im_{1\infty} = \Im_{14}$.\\

$\Im_{NMI}^2  = \{a:[0.24,0.656], g:[0.06,0.5866]\}$.; and $||D_a||_\infty = max\{0.36,0.144\} = 0.36 > \epsilon$ and $||D_g||_\infty = max\{0.06,0.1134\} = 0.1134 > \epsilon$

As NMI iteration proceeds, the epistemic states of the chosen elements are modified as follows:

$\Im_{NMI}^3 = \{a:[0.46464,0.72063], g:[0.11501,0.63749]\}$;

$\Im_{NMI}^4 = \{a:[0.35328,0.66525], g:[0.10868,0.59389]\}$;

$\Im_{NMI}^5 = \{a:[0.41107,0.68522], g:[0.12211,0.60961]\}$;

$\Im_{NMI}^6 = \{a:[0.38348,0.67162], g:[0.11954,0.5989]\}$;

$\Im_{NMI}^7 = \{a:[0.39742,0.67695], g:[0.1226,0.6031]\}$;

$\Im_{NMI}^8 = \{a:[0.39078,0.67381], g:[0.12181,0.60063]\}$;

The iteration is terminated here. The \textit{termination of the iterations} is decided solely by the intended accuracy of computation. The choice of $\epsilon = 0.009$ ensures accuracy up to $2^{nd}$ decimal point. The monotonic iteration of $\Gamma$ on $P_{Ex.7}(\Im_{NMI}^8)$ which gives the epistemic states of all other atoms proceeds as follows:

$\Im_{1} = \Gamma_{P_{Ex.7}(\Im_{NMI}^8)}(\Phi) = \{b:[0.65682, 0.84393], d:[0.15607,0.59173]\}$;

$\Im_{2} = \Gamma_{P_{Ex.7}(\Im_{NMI}^8)}(\Im_{1}) = \Im_{1} \cup \{c:[0.65682,0.84393], e:[0.140463,0.59173]\}$;

$\Im_{3} = \Gamma_{P_{Ex.7}(\Im_{NMI}^8)}(\Im_{2}) = \Im_{2} \cup \{a:[0.39409,0.67514], f:[0.40827,0.85954]\}$;

$\Im_{4} = \Gamma_{P_{Ex.7}(\Im_{NMI}^8)}(\Im_{3}) = \Im_{3} \cup \{g:[0.12248,0.60168]\}$;

$\Im_{\infty} = \Im_{4}$.

Therefore, considering up to $2^{nd}$ decimal place and rounding off the values the solution becomes $\{a:[0.39,0.67], b:[0.66,0.84], c:[0.66,0.84], d:[0.16,0.52], e:[0.14,0.52], f:[0.48,0.86], g:[0.12,0.60]\}$, which is the intended answer set.\\

Thus NMI iteration comprises of several stages of monotonic iterations using the immediate consequence operator $\Gamma$.\\

\textbf{Mathematical analysis of NMI iterations:}

A proper mathematical model is needed to study the convergence of iterations described in previous subsection. The assumption set for a particular program segment (represented by an SCC) gives a way to "unfurl" the interconnected cycles into linear paths along which the computation progresses and the epistemic states of atoms in the assumption set propagates from one iteration step to the next one. The value propagation graph enables us to describe the NMI iteration process by means of a system of difference equations in terms of the atoms in $As_P$.

\textbf{Definition 13:}

A \textit{difference equation} of order $(k+1)$ is an equation of the form

\begin{center}

$x_{n+1} = f(x_n,x_{n-1},...,x_{n-k}),     n = 0,1,...$

\end{center}

where $f$ is a continuous function from $D^{k+1} \rightarrow D$, for some domain D.

\textbf{Example 7 (contd)} From the value propagation graph shown in Figure \ref{figure:second example propagation} the following equations can be written.

$[a_{1_n},a_{2_n}] = [0.6,0.8]\wedge[c_{1_n},c_{2_n}] = [0.6c_{1_n}, 0.8c_{2_n}]$;

$[c_{1_n},c_{2_n}] = [b_{1_n},b_{2_n}];$

$[b_{1_n},b_{2_n}] = not \ [a_{1_{(n-1)}},a_{2_{(n-1)}}] \vee [g_{1_{(n-1)}},g_{2_{(n-1)}}];$

$ \  \  \  \  \  \  \  \  \ = [1-a_{1_{(n-1)}}, 1-a_{1_{(n-1)}}] \vee [g_{1_{(n-1)}},g_{2_{(n-1)}}];$

$ \  \  \  \  \  \ \   \  \ = [1-a_{1_{(n-1)}}+a_{1_{(n-1)}}.g_{1_{(n-1)}},1-a_{1_{(n-1)}}+a_{1_{(n-1)}}.g_{2_{(n-1)}}]$

$[g_{1_n},g_{2_n}] = [0.3,0.7] \wedge [f_{1_n},f_{2_n}] = [0.3f_{1_n}, 0.7f_{2_n}]$;

$[f_{1_n},f_{2_n}] = \neg [e_{1_n},e_{2_n}] = [1-e_{2_n}, 1-e_{1_n}];$

$[e_{1_n},e_{2_n}] = [0.9, 1] \wedge [d_{1_n},d_{2_n}] = [0.9d_{1_n}, d_{2_n}]$;

$[d_{1_n},d_{2_n}] = [a_{1_{(n-1)}},a_{2_{(n-1)}}] \wedge (\neg [g_{1_{(n-1)}},g_{2_{(n-1)}}])$

$ \  \  \ \  \  \  \  \  \  \  \ = [a_{1_{(n-1)}}.(1-g_{2_{(n-1)}}),a_{2_{(n-1)}}.(1-g_{1_{(n-1)}})]$.

Thus,

$[g_{1_n},g_{2_n}] = [0.3 - 0.3 \times 1 \times d_{2_n}, 0.7 - 0.7 \times 0.9 \times d_{1_n}]$

$ \  \  \  \  \  \  \  \  \  \  \  \ = [0.3 - 0.3a_{2_{(n-1)}}.(1-g_{1_{(n-1)}}), 0.7 - 0.54a_{1_{(n-1)}}.(1-g_{2_{(n-1)}})]$.

Hence,

$a_{1_n} = 0.6 - 0.6a_{1_{(n-1)}} \cdot (1-g_{1_{(n-1)}})$

$a_{2_n} = 0.8 - 0.8a_{1_{(n-1)}} \cdot (1-g_{2_{(n-1)}})$

$g_{1_n} = 0.3 - 0.3a_{2_{(n-1)}} \cdot (1-g_{1_{(n-1)}})$

$g_{2_n} = 0.7 - 0.54a_{1_{(n-1)}} \cdot (1-g_{2_{(n-1)}})$.

Together, the above equations can be expressed as\\

\[
  \bar{\textbf{a}}^n =
  \begin{bmatrix}
           a_{1_n} \\
           a_{2_n} \\
           g_{1_n}\\
					 g_{2_n}
  \end{bmatrix} =
  \begin{bmatrix}
           f_{a_1}(a_{1_{(n-1)}}, a_{2_{(n-1)}}, g_{1_{(n-1)}},g_{2_{(n-1)}})\\
           f_{a_2}(a_{1_{(n-1)}}, a_{2_{(n-1)}}, g_{1_{(n-1)}},g_{2_{(n-1)}})\\
           f_{g_1}(a_{1_{(n-1)}}, a_{2_{(n-1)}}, g_{1_{(n-1)}},g_{2_{(n-1)}})\\
					 f_{g_2}(a_{1_{(n-1)}}, a_{2_{(n-1)}}, g_{1_{(n-1)}},g_{2_{(n-1)}})
  \end{bmatrix} = \bar{\textbf{F}}(\bar{\textbf{a}}^{n-1}).
\]

Hence, NMI iterations can be described by a \textit{system of first order nonlinear difference equations}. Termination of this NMI iteration means that the iteration has reached to a point $\bar{\textbf{x}} \in [0,1]^4$ such that 

\begin{center}
$\bar{\textbf{x}} = \bar{\textbf{F}}(\bar{\textbf{x}}).$
\end{center}

Thus, the problem of studying the convergence of the NMI iteration reduces to the problem of finding the iterative fixed-point for the system of difference equations. 

On a generalized setting, suppose for some program $P$, $As_P = \{a_1,a_2,...,a_m\}$, then $\bar{\textbf{a}} = [a_{1_1} \  \  a_{1_2} \  \ ... \  \  a_{m_1} \ \ a_{m_2}]^T$  and $\bar{\textbf{F}}$ is a mapping from $[0,1]^{2m} \rightarrow [0,1]^{2m}$, given by $\bar{\textbf{F}} = [f_1(\bar{\textbf{a}}) \ f_2(\bar{\textbf{a}}) \ ... \ f_{2m}(\bar{\textbf{a}})]^T$ and $\bar{\textbf{a}}^{n+1} = \bar{\textbf{F}}(\bar{\textbf{a}}^n)$.

The mapping $\bar{\textbf{F}}$, which is the set of functions describing the iteration, is called the \textbf{iteration function} of P and $f_1, f_2, ..., f_{2m}$ are the component functions of $\bar{\textbf{F}}$.

One point should be noted that, the iteration function, being dependent on the assumption set, is not unique for an SCC. Therefore, proper selection of assumption set is important in order to attain some well-behaved iteration function that is convergent.

\textbf{Conditions for convergence:}

Sufficient conditions for convergence of the iterative functions for any starting epistemic states are investigated in this subsection.

\textbf{1. Programs without \textit{not} and $\otimes_k$:}

\textbf{Lemma 1}

If a program segment $P$ doesn't contain negation-as-failure and the knowledge aggregation operator $(\otimes_k)$, then for any atom $a\in AS_P$ and for any $n >0$ at the $n^{th}$ and $(n+1)^{st}$ steps of NMI iteration $a_{1_n} \leq a_{1_{(n+1)}} \leq a_{2_{(n+1)}} \leq a_{2_n}$, with $a_0 = [a_{1_0},a_{2_0}] = [0,1]$, where $[a_{1_i}, a_{2_i}]$ is epistemic state of $a$ at the $i^{th}$ iteration.

\textbf{Proof:} Proof is presented in the Appendix section.

\textbf{Theorem 4}

If a program segment doesn't contain negation-as-failure and the knowledge aggregation operator $(\otimes_k)$ then NMI iteration, started from the initial epistemic state $[0,1]$, terminates and gives the answer set. 

\textbf{Proof:} Proof is presented in the Appendix section.

\textbf{2. Program segment whose dependency graph is a simple cycle without any $\otimes_k$:}

\textbf{Definition 14:} Suppose the dependency graph of a program segment there is a simple cycle $S$ connected with some constant intervals from $\mathscr{T}$ as conjuncts and disjuncts. Then the gain of the cycle as observed from a particular node $a \in \textbf{B}_S$ (i.e. $a$ is in the assumption set) is given by $\lVert \mathbf{G}_{S} \rVert_\infty$, where $\mathbf{G}_{S} = [G_1 \ G_2]$, such that at any iteration $[a_{1_n}, a_{2_n}] = [X + G_1 \times a_{i_{(n-1)}}, Y + G_2 \times a_{j_{(n-1)}}]$, where, $X,Y$ are constants and $i,j \in \{1,2\}$, depending on the $\neg$ or $not$ present in the cycle.

Consider a simple cycle with some atom $a$ as the chosen element. Say, there are N nodes, $n_1$, $n_2$, $... n_N$, in the simple cycle apart from the node containing $a$. The set of nodes is expanded to $n_0$, $n_1$, $n_2$, $... n_N$, $n_{N+1}$, where, $n_0 = n_{N+1} = a$. The intervals $[x_1,y_1], [x_2,y_2],...,[x_n,y_n]$ from $\mathscr{T}$ are connected to the conjunctive ($\wedge$) nodes of the simple cycle and $[u_1,v_1], [u_2,v_2],...,[u_m,v_m]$ from $\mathscr{T}$ are connected to the disjunctive ($\vee$) nodes of the simple cycle. The procedure for calculating the gain is described in Algorithm 1.

\begin{figure}
\centering
\begin {tikzpicture}[-latex ,auto ,node distance =1.0 cm and 1.2 cm ,on grid ,
semithick ,
state/.style ={ circle,
draw, text=black , minimum width = 0.5 cm}]

\node (Ai){$a_m$};
\node[state] (Con1)[right =of Ai]{$\wedge$};
\node (x1)[below =of Con1]{$[x_{1},y_{1}]$};
\node[state] (Con2)[right =of Con1]{$\wedge$};
\node (x2)[below =of Con2]{$[x_{2},y_{2}]$};
\node[state] (Dis1)[right =of Con2]{$\vee$};
\node (x3)[below =of Dis1]{$[x_{3},y_{3}]$};
\node[state] (Con3)[right =of Dis1]{$\wedge$};
\node (x4)[below =of Con3]{$[x_{4},y_{4}]$};
\node (Ao)[right =of Con3]{$a_{m+1}$};

\path (Ai) edge (Con1);
\path (Con1) edge node[above =0.15 cm]{$\neg$} (Con2);
\path (Con2) edge (Dis1);
\path (Dis1) edge node[above =0.15 cm]{$-1$}(Con3);
\path (Con3) edge (Ao);
\path (x1) edge (Con1);
\path (x2) edge (Con2);
\path (x3) edge (Dis1);
\path (x4) edge (Con3);

\end{tikzpicture}
\caption{Example for Gain Calculation}
\label{figure:GainEx}
\end{figure}
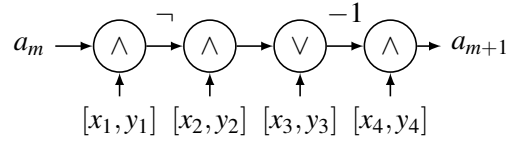

\textbf{Example 8:} Consider a program segment whose corresponding dependency graph is a simple cycle and for any chosen atom $a$ the corresponding value propagation graph is as shown in figure \ref{figure:GainEx}. Now,

$[a_{1_{(m+1)}}, a_{2_{(m+1)}}] = [x_4(1-x_3)(1-x_2)+x_4x_2y_1(1-x_3)a_{2_m}, y_4(1-x_3)(1-x_2)+x_4x_2y_1(1-x_3)a_{1_m}]$

Following Algorithm 1 we have $G = [y_1x_2x_4(1-x_3) \  \ y_1x_2y_4(1-x_3)]$; and the Gain is $\lVert G \rVert_{\infty} = max(\lvert y_1x_2x_4(1-x_3)\rvert, \lvert y_1x_2y_4(1-x_3)\rvert)$.

\textbf{Definition 15:}\cite{ortega2000iterative} A mapping $\bar{\textbf{F}}:[0,1]^n \longrightarrow [0,1]^n$ is \textit{nonexpansive} on $[0,1]^n$ if for any $\bar{x},\bar{y} \in [0,1]^n$

\begin{center}

$\lVert \bar{\textbf{F}}(\bar{x}) - \bar{\textbf{F}}(\bar{y}) \rVert \leq \lVert x-y \rVert.$

\end{center}

A mapping $\bar{\textbf{F}}:[0,1]^n \longrightarrow [0,1]^n$ is a \textit{contraction mapping} on $[0,1]^n$ (or simply a \textit{contraction}) if there is a $\alpha < 1$ such that $\lVert \bar{\textbf{F}}(\bar{x}) - \bar{\textbf{F}}(\bar{y}) \rVert \leq \alpha \lVert x-y \rVert$ for any $\bar{x},\bar{y} \in [0,1]^n$. Such a $\alpha$ is called a \textit{contraction modulus} or \textit{Lipscitz Constant}.

\textbf{Theorem 5 (Contraction mapping theorem:)} \cite{ortega2000iterative} Let $\bar{\textbf{F}}:D \subset R^n \longrightarrow R^n$ maps a closed set $D_0 \subset D$ into itself and that $\lVert \bar{\textbf{F}}(\bar{x}) - \bar{\textbf{F}}(\bar{y}) \rVert \leq \alpha \lVert \bar{x} - \bar{y} \rVert$ for any $\bar{x},\bar{y} \in D_0$ for some $\alpha < 1$. Then for any $\bar{x}^0 \in D_0$, the sequence $\bar{x}^{k+1} = \bar{\textbf{F}}(\bar{x}^k)$ for $k = 0,1,..$, converges to the unique fixed point $\bar{x}^*$ of $\bar{\textbf{F}}$ in $D_0$.

\begin{algorithm}[H]
\SetAlgoLined
\hrule
\hrule
 initialization: flag = 0, $G_1 = 1, G_2 = 0, G = 0$\;
 \For {$i \gets 1$ to $N$}{
  \uIf{$w(e(n_i,n_{i-1})) == \neg \  \ \text{and}$ \ \text{flag} == 1}{
    $G_1 \leftrightarrow G_2$ \;
  }
  \uElseIf{$w(e(n_i,n_{i-1})) == -1 \  \ \text{and}$ \ \text{flag} == 1}{
    $G_2 \gets G_1$ \;
  }
  \Else{
    Do nothing;
  }
	
	  \uIf{$n_i = \wedge \  \ \text{and}$ \ $e(n_i,[x_i,y_i]) \in E$}{
    $\text{flag} \gets 1$ \;
		$G_1 \gets G_1 \times x_i$ \;
		$G_2 \gets G_2 \times y_i$ \;
  }
  \uElseIf{$n_i = \vee \  \ \text{and}$ \ $e(n_i,[u_i,v_i]) \in E$}{
    $G_1 \gets G_1 \times (1-u_i)$ \;
		$G_2 \gets G_2 \times (1-v_i)$ \;
  }
  \Else{
    Do nothing;
  }
 }
$G = max (\lvert G_1 \rvert, \lvert G_2 \rvert)$\;
 \caption{Calculating Gain of a Simple Cycle}
\end{algorithm}
\hrule
\hrule

The following theorem relates the constant $\alpha$ of Theorem 5 to the partial derivatives of the components of the iteration function.

\textbf{Theorem 6:} \cite{atkinson2008introduction} Let $\bar{\textbf{F}}:D \subset R^n \longrightarrow R^n$ maps a closed set $D_0 \subset D$ into itself and the components of $\bar{\textbf{F}}$ are continuously differentiable at all points of $D_0$ and further assume $\max\limits_{\bar{x} \in D_0} \lVert J_{\bar{F}}(\bar{x}) \rVert_{\infty} < 1$; where, $\bar{\textbf{F}} = [f_1,f_2,...,f_n]^T$ and $J_{\bar{F}}(\bar{x})$ is the $n \times n$ Jacobian matrix at some point $\bar{x} \in D_0$ with element $J_{\bar{F}}(\bar{x})_{ij} = \frac{\partial f_i(\bar{x})}{\partial x_j}, i,j = 1,...,n$. Then,\\

1. $\bar{x} = \bar{\textbf{F}}(\bar{x})$ has a unique solution $\bar{\alpha} \in D_0$.

2. For any initial point $\bar{x_0} \in D_0$, the iteration $\bar{x}_k = \bar{\textbf{F}}(\bar{x}_{k-1}), k =1,2,...$ will converge in $D_0$ to $\bar{\alpha}$.

\textbf{Theorem 7:} Let $P$ be a transformed program, whose dependency graph contains exactly one simple cycle, such that none of its nodes is $\otimes_k$ and all the conjuncts and disjuncts to the simple cycle are constant intervals from $\mathscr{T}$. The NMI iteration for such a program converges, for any initial epistemic state of the chosen elements, if the gain of the cycle is $ < 1$.
 
\textbf{Proof:} Proof is presented in the Appendix section.

\textbf{3. Program segment with 'not', whose dependency graph is a SCC with multiple simple cycles without any $\otimes_k$:}

Consider for some SCC corresponding to some program segment, $m$ atoms are chosen to construct the assumption set and $As_{P_{SCC}} = \{a_1, a_2,...,a_m\}$. The iteration function $\bar{F}$ has $n$ component functions, where $n = 2m$. Say, $\bar{F} = [f_1 \  f_2 \ ... \ f_n]^T$, where, $a_i = [a_{i,1},a_{i,2}]$ and $f_1$ corresponds to $a_{1,1}$ and $f_2$ corresponds to $a_{1,2}$ so on. Each row of the Jacobian matrix for $\bar{F}$ is of the form

\begin{center}

$J_{\bar{F}}(\bar{x})_i = [\frac{\partial f_i(\bar{x})}{\partial a_{1,1}} \ \frac{\partial f_i(\bar{x})}{\partial a_{1,2}} \cdot \cdot \cdot \frac{\partial f_i(\bar{x})}{\partial a_{m,1}} \  \ \frac{\partial f_i(\bar{x})}{\partial a_{m,2}}]$

\end{center}

for some point $\bar{x} \in [0,1]^n$.

So, satisfaction of the sufficient condition for convergence as specified in Theorem 6, requires,

\begin{center}

$\max\limits_{\bar{x} \in D_0} (\max\limits_{i \in \{1,2,..,n\}} \lVert J_{\bar{F}}(\bar{x})_i \rVert_{\infty}) < 1$

\end{center}

i.e., for every $1 \leq i \leq n$, $\lVert J_{\bar{F}}(\bar{x})_i \rVert_{\infty} < 1$ for all $\bar{x} \in [0,1]^n$. From the definition of vector norms,
\begin{center}
$\lVert J_{\bar{F}}(\bar{x})_i \rVert_{\infty} = \lvert \frac{\partial f_i(\bar{x})}{\partial a_{1,1}} \rvert + \lvert \frac{\partial f_i(\bar{x})}{\partial a_{1,2}} \rvert + \lvert \frac{\partial f_i(\bar{x})}{\partial a_{m,1}} \rvert + \lvert \frac{\partial f_i(\bar{x})}{\partial a_{m,2}} \rvert$.
\end{center}

Increasing the number of chosen element would increase the number of independent variables in each of the component functions of the iteration function, thus inserting additional terms in the rows of the Jacobian matrix. The incorporation of additional terms makes it more difficult to satisfy the sufficient condition for being contraction. Thus, \textit{keeping the number of chosen elements as small as possible is intended}.

\textbf{Lemma 2} Suppose $f:[0.1]^m \longrightarrow [0,1]$ be a logical function (constructed using product t-norm, t-conorm and negation operators only) of $m$ variables, say $0 \leq a_1,a_2,...,a_m \leq 1$, for any $m \geq 2$. Then

\begin{center}

$\lvert \frac{\partial f}{\partial a_1}\rvert + \lvert \frac{\partial f}{\partial a_2}\rvert + ... + \lvert \frac{\partial f}{\partial a_m}\rvert \leq m$.

\end{center}

\textbf{Proof:} Proof is presented in the Appendix section.

\textbf{Theorem 8}

Suppose the value-propagation path for some chosen element $a$, has only conjunctive nodes ($\wedge$), and no disjunctions and moreover, among the conjuncts, $k$ conjuncts are not constants but come from value-propagation paths of other chosen elements or from other nodes of the value-propagation path of $a$. These $k$ conjuncts, say $g_1,g_2,..,g_k$, vary over iterations. Then the infinite norm ($\lVert \ \rVert_\infty$) of the rows corresponding to $a$ in the Jacobian matrix will be $< 1$ if the path gain of the vpp of $a$(considering only the constant conjuncts) is strictly less that $\frac{1}{k+2}$.

\textbf{Proof:} Proof is presented in the Appendix section.

If the vpp contains disjunctions then satisfaction of the sufficient condition depends on the chosen assumption set and cannot be controlled by the path gain alone. This is illustrated with the help of the following examples.

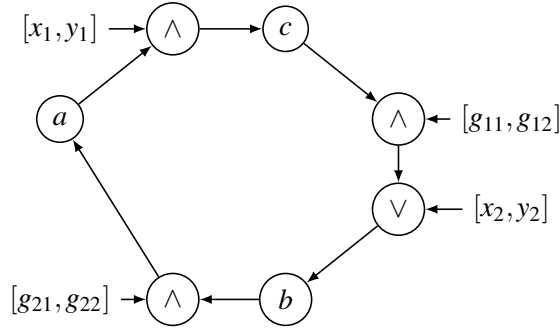
\begin{figure}
\centering
\begin {tikzpicture}[-latex ,auto ,node distance =1.2 cm and 1.5cm ,on grid ,
semithick ,
state/.style ={ circle,
draw, text=black , minimum width = 0.5 cm}]

\node[state] (A){$a$};
\node[state] (Con1)[above right =of A]{$\wedge$};
\node[state] (C)[right =of Con1]{$c$};
\node[state] (Con2)[below right =of C]{$\wedge$};
\node[state] (Dis)[below =of Con2]{$\vee$};
\node[state] (B)[below left =of Dis]{$b$};
\node[state] (Con3)[left =of B]{$\wedge$};
\node (nc) [left =of Con1]{$[x_1,y_1]$};
\node (G1) [right =of Con2]{$[g_{11},g_{12}]$};
\node (nb) [right =of Dis]{$[x_2,y_2]$};
\node (G2) [left =of Con3]{$[g_{21},g_{22}]$};

\path (A) edge (Con1);
\path (Con1) edge (C);
\path (C) edge (Con2);
\path (Con2) edge (Dis);
\path (Dis) edge (B);
\path (B) edge (Con3);
\path (Con3) edge (A);
\path (nc) edge (Con1);
\path (G1) edge (Con2);
\path (nb) edge (Dis);
\path (G2) edge (Con3);

\end{tikzpicture}
\caption{Value propagation path containing disjunction}
\label{figure:example_disjunction}
\end{figure}

 Consider the section of an SCC as shown in Figure \ref{figure:example_disjunction}, where, $[g_{11},g_{12}]$ and $[g_{21},g_{22}]$ are inputs coming from other sections of the SCC and they vary with iterations. Choosing $a$, $b$ or $c$ in assumption set gives rise to different iteration functions, as shown here:

$a_{n+1} = [g_{21}(x_2 + (1 - x_2)x_1a_{n,1}g_{11}), g_{22}(y_2 + (1-y_2)y_1a_{n,2}g_{12})]$;

$b_{n+1} = [x_2 + (1-x_2)x_1b_{n,1}g_{21}g_{11}, y_2 + (1-y_2)y_1b_{n,1}g_{22}g_{12}]$;

$c_{n+1} = [x_1x_2g_{21} + x_1(1-x_2)c_{n,1}g_{11}g_{21}, y_1y_2g_{22} + y_1(1-y_2)c_{n,2}g_{12}g_{22}]$.

The infinite norm of the corresponding rows of the Jacobian matrix are given by;

$\lVert J_a \rVert = max(x_2 + x_1(1-x_2)(g_{11}g_{21}+a_{n,1}g_{21}+a_{n,1}g_{11}), y_2 + y_1(1-y_2)(g_{12}g_{22}+a_{n,2}g_{22}+a_{n,2}g_{12})) \leq max(x_2 + 3x_1(1-x_2), y_2 + 3y_1(1-y_2))$.

$\lVert J_b \rVert = max(x_1(1-x_2)(g_{11}g_{21}+b_{n,1}g_{21}+b_{n,1}g_{11}), y_1(1-y_2)(g_{12}g_{22}+b_{n,2}g_{22}+b_{n,2}g_{12})) \leq max(3x_1(1-x_2), 3y_1(1-y_2))$.

$\lVert J_c \rVert = max(x_1x_2 + x_1(1-x_2)(g_{11}g_{21}+c_{n,1}g_{21}+c_{n,1}g_{11}), y_1y_2 + y_1(1-y_2)(g_{12}g_{22}+c_{n,2}g_{22}+c_{n,2}g_{12})) \leq max(x_1x_2 + 3x_1(1-x_2), y_1y_2 + 3y_1(1-y_2))$.

Clearly atom $b$ is the best choice for assumption set, since in this case the sufficient condition can be satisfied only by making the path gain sufficiently low.

\textbf{Note:} Since, presence of disjunction in the vpp gives rise to extra terms in the differential expression in the Jacobian matrix, choosing the atom that is the predecessor of the disjunctive node (since rules are represented in DNF it is guaranteed that there will be an atom) is better choice, as can be seen form figure \ref{figure:example_disjunction}. This imposes another condition for constructing assumption set along with the general conditions stated in section 5.4.1.

\textbf{4.Program segments with $not$ and $\otimes_k$}
	
The convergence of NMI iteration on programs with $\otimes_k$ can not be guaranteed in general. Even when a program segment has a unique answer set, the plain NMI iteration may not reach to it. The following example illustrates this fact.

\textbf{Example 8:} Consider a transformed program segment:

$P_{Ex.8} = \{r_1^*: b \longleftarrow a \wedge [0.3,0.5]; \  \  \ r_2^*: c \longleftarrow \neg b \otimes_k [0.4,0.9]; \  \  \ r_3^*: a \longleftarrow \neg c \}$.

Say $As_{P_{Ex.8}} = \{a\}$. Now the iteration proceeds as follows:

1. $\Im_{NMI}^0 = \{a:[0,1]\}$;

2. $\Im_{NMI}^1 = \{a:[0.5,1] \otimes_k [0.4,0.9]\}$;

$ \  \  \  \  \  \  \  \  \  \  \  \ \{a:[\xi,\xi]\}$.

And the NMI iteration halts here. But the program segment $P_{Ex.8}$ has a unique answer set; $\{a:[0,0], b:[0,0], c:[1,1]\}$, which is not attained by NMI iteration.

	\textbf{Theorem 9}

Consider a set of transformed rules $P_c$, such that,

$\bullet$ The dependency graph of $P_c$ forms a simple cycle

$\bullet$ $P_c$ has only one rule involving $\otimes_k$, which is of the form

\begin{center}

$r: a \longleftarrow \bar{c} \otimes_k B_1$;

\end{center}

where, $\bar{c} \in \mathscr{T}$ and $B_1$ is an expression in CNF or DNF form involving
 literals from $Lit_{P_c}$. All the other rules of $P_c$ do not contain $\otimes_k$.

$\bullet$ Let $P_c^{-\otimes}$ is the transformed program:

\begin{center}

$\{P_c \setminus r\} \cup \{a \longleftarrow\ B_1\}$

\end{center}

and the iteration function of $P_c^{-\otimes}$, for any chosen element, is a contraction mapping, i.e., it has a unique fixed point and NMI iteration converges to that fixed point ($\Im^{-\otimes}$). Let $v_{\Im^{-\otimes}}(B_1)$ denotes the epistemic state of $B_1$ obtained from $\Im^{-\otimes}$.

$\bullet$ Let a single NMI iteration with assumption set $As_{P_c} = \{a\}$ and initial interpretation $\Im_s = \{a:\bar{c}\}$ assigns the epistemic state $v_{\Im_s}(B_1)$ to $B_1$.

Then it can said:

1. If $v_{\Im^{-\otimes}}(B_1) >_{k_p} \bar{c}$, then $\Im^{-\otimes}$ is a supported model of $P_c$ with $\Im^{-\otimes}(a) = v_{\Im^{-\otimes}}(B_1)$.

2. If $v_{\Im_s}(B_1) <_{k_p} \bar{c}$, then $\Im_s$ is a supported model of $P_c$ with $\Im_s(a) = \bar{c}$.

3. If $\Im^{-\otimes} \leq_{k_p} \Im_s$. then $\Im^{-\otimes}$ is the unique answer set of $P_c$.

4. If $\Im_s \leq_{k_p} \Im^{-\otimes}$, then $\Im_s$ is the unique answer set of $P_c$.

5. If $\Im_s \nleq_{k_p} \Im^{-\otimes}$ and $\Im^{-\otimes} \nleq_{k_p} \Im_s$, then both are answer sets of $P_c$.

\textbf{Proof:} Proof is presented in the Appendix section.

Following Theorem 9, Example 8 can be analyzed. The iteration function for $P_{Ex.8}^{-\otimes}$ is a contraction mapping, giving rise to the unique fixed point $\Im^{-\otimes} = \{a:[0,0], b:[0,0], c:[1,1]\}$ and $v_{\Im^{-\otimes}}(\neg b) = [1,1] >_{k_p} [0.4,0.9]$. So this is an answer set of $P_{Ex.8}$. If NMI iteration is performed with initial interpretation $\Im_s = \{c:[0.4,0.9]\}$ the resulting interpretation becomes $\Im_s = \{a:[0.1,0.6], b:[0.03,0.3]\}$. So $v_{\Im_s}(\neg b) = [0.7,0.97] >_{k_p} [0.4,0.9]$. Thus the NMI iteration does not give a stable valuation. Hence, the unique answer set is $\Im^{-\otimes}$ as was mentioned in the example.

The assumption that one input to the $\otimes_k$ is a constant from $\mathscr{T}$ is not very unintuitive. It ensures that, the two subprograms, on which the two inputs of $\otimes_k$ depend, are independent and don't have any common literal. This signifies that positive and negative evidences for any atom don't share any common literal and this assumption is quite natural.

If the simple cycle in Theorem 9 is replaced by a SCC, then, following same steps of Theorem 9, stable and supported valuations can be determined. But to decide whether they are answer sets or not, requires to consider all possible assumption sets from the SCC and all possible stable NMI iterations. So this case is not considered here. 

\textbf{Note:} 1. The conditions stated in the above theorems are sufficient conditions and not necessary conditions. Thus, meeting those conditions guarantees termination to a unique fixed point. But if those conditions are not satisfied then also the NMI iteration may terminate to  unique fixed point, as is clear from the Example 7.

2. It is assumed that for a program segment with $not$ the values of the constant conjuncts and disjuncts are such that NMI iterations always terminate at a unique fixed point.
 
\subsubsection{Branch and Bound}

For simple cycles with unity gain or SCCs with no constant conjunct or disjunct, there may be more than one stable valuations, not attained by the NMI iterations.

For example, for the set of rules $P = \{a \longleftarrow not \ b, b \longleftarrow not \ a\}$, with $As_P = \{a\}$, any valuation of the form $\{a:[x,x]| x \in [0,1]\}$ is a stable valuation and the corresponding answer set is $\{a:[x,x], b:[1-x,1-x]\}$. But NMI iteration, would terminate at a value dependent on the starting epistemic state. All other answer sets cannot be obtained. In such a scenario, branch-and-bound is required.

Firstly assumption set is constructed from the SCC following the same conditions specified in section 5.4.1, with addition to the following criteria:\\

The chosen atom has to be at the head of an edge having weight '-1', i.e., the chosen element, say $a$, occurs as $a \longleftarrow not  \ b$ for some $b$ in the program segment.\\

Every possible epistemic state of the form $[x,x]$ can not be checked for stability for all $x \in [0,1]$. Therefore, a predefined positive integer, $N_B$, is chosen, so that at any time the branch-and-bound is called for, only $N_B$ equidistant points are are chosen from $[0,1]$ and only those epistemic states are checked for stability. For all the elements in the assumption set all combinations of $N_B$ epistemic states are checked for stability.

After assigning each set of values monotonic iteration is performed.

Say, $S_1$ be an SCC to be evaluated using branch-and-bound. An atom, say $a$, is connected to a conjunction node of SCC $S_2$, placed just higher position in the topologically sorted list of SCCs. Thus the course of iteration and the epistemic states of atoms in $S_2$ depend upon the epistemic state of atom $a$ in $S_1$. So for each stable valuation of $a$, a version of $S_2$ is generated with that value of $a$ coming to the conjunctive node of $S_2$. Maximum $N_B$ versions of $S_2$ can be generated if each of the assignments gives stable valuation of $a$. In other words, whenever an SCC is evaluated using branch-and-bound the computation of the next SCCs proceeds in different branches.

Here, it is \textit{assumed} $S_2$ satisfies the sufficient condition for being a contraction even if the contribution of $a$ is absent. 

Therefore, the Nonmonotonic Evaluation is not \textit{semantically complete}, in the sense that all possible answer sets can not be obtained.

\subsubsection{Nonmonotonic Evaluation Procedure:}

From the topologically sorted list of SCCs, obtained from the dependency graph of the reduced program after MI stage, one after another is chosen along with the associated constant nodes.

1. If the SCC has no $\otimes_k$ nodes and some constant intervals are connected to some conjunctive ($\wedge$) and disjunctive ($\vee$) nodes, then NMI iteration is called. It is assumed here, that the constant values are such that with proper selection of assumption set the sufficient condition for convergence of iteration is met and the unique fixed point is reached through iteration.

2. If the SCC has no constant interval connected to any of the conjunctive ($\wedge$) and disjunctive ($\vee$) nodes and no edge is of weight $-1$ and no node is $\otimes_k$, then also NMI iteration started. In this case, all the relevant atoms get $[0,1]$ as their epistemic state.

3. If the SCC is a simple cycle with one $\otimes_k$ node and one input to the $\otimes_k$ is an element $\bar{c} \in \mathscr{T}$ then it can have at most two answer sets, which can be obtained following Theorem 9.

4. If the SCC has no constant interval connected to some conjunctive ($\wedge$) and disjunctive ($\vee$) nodes and at least one edge of weight $-1$ is present, branch-and-bound is called for.

5. With evaluation of each SCC, the SCC placed just higher position in the topologically sorted list gets modified by the evaluated epistemic states of the atoms in the former one.

\textbf{Example 6(contd):}

From topologically sorted list of SCCs, $hijk, \ uvxw, \ c, \ gbadef, \ yz, \ l$, one after one is chosen and epistemic states are evaluated.

1. $SCC_{hijk}$ is chosen. $As_{hijk} = \{h\}$.

Iteration function for $SCC_{hijk}^{-\otimes_k}$ is a contraction mapping (Theorem 7) and for any initial value NMI iteration terminates at a unique epistemic state of atom $h$, which is $h:[0.5557, 0.7938]$. From this epistemic states of other atoms of $SCC_{hijk}^{-\otimes_k}$ are calculated as:

$v(hijk) = \{h:[0.5557, 0.7938], i:[0.4443, 0.4443], j:[0.2062, 0.2062], k:[0.7938, 0.7938]\}$.

Now, in $SCC_{hijk}$ $[0.44,0.58] \leq_{k_p} [0.2062,0.2062]$; hence $v(hijk)$ gives the answer set of the atoms in $SCC_{hijk}$ (Theorem 9). This is the unique answer set since the NMI iteration with starting point $\{j:[.44, 0.58]\}$  does not offer a stable valuation.
 
2. The epistemic state of atom $c$ is $\{c:[0.5557, 0.7938]\} = v(h)$. 

3. $SCC_{uvxw}$ is chosen. $As_{uvxw} = \{x\}$.

NMI iterations gives:

$\{x:[0.16212,0.28255], w:[0.16212, 0.28255], u:[0.0811, 0.22604], v:[0.8106, 0.9418]\}$.

3. $SCC_{gbadef}$ is modified by already calculated epistemic states of atom c and w.

$As_{gbadef} = \{a,g\}$.

NMI iteration gives: $\{a:[0.2967,0.4116], g:[0.1834, 0.4322], b:[0.7577, 0.8315], d:[0.1685, 0.3361], e:[0.0271, 0.0949], f:[0.9050, 0.9727]\}$.

4. In $SCC_{yz}$, since there is no conjuncts or disjuncts, branch-and-bound evaluation is followed. Say $N_B$ is chosen to be 4 and $As_{yz} = \{y\}$.

So, Monotonic iterations are done with four initial values, i.e., $y:[0,0]$, $y:[0.25,0.25]$, $y:[0.75,0.75]$ and $y:[1,1]$. All of them are stable; hence we get four answer sets respectively

$v_1 = \{y:[0,0], z:[1,1]\}$, $v_2 = \{y:[0.25,0.25], z:[0.75,0.75]\}$, $v_3 = \{y:[0.75,0.75], z:[0.25,0.25]\}$ and $v_4 = \{y:[1,1], z:[0,0]\}$.

5. Based on the epistemic states of $z$, $l$ has four possible values corresponding to each of the values of $z$. They are $v_1 = \{l:[0.4,0.6]\}$, $v_2 = \{l:[0.3,0.45]\}$, $v_3 = \{l:[0.1,0.15]\}$ and $v_4 = \{l:[0,0]\}$.

Therefore, for the UnASP program in Example 7 the four computed answer sets are as follows:

Suppose $v = \{q:[0.7,0.7], r:[1,1], n:[0.7,0.9], t:[0,1], m:[0.42,0.56], s:[0.42,0.56], p:[0.3916,0.495], h:[0.5557, 0.7938], i:[0.4443, 0.4443], j:[0.2062, 0.2062], k:[0.7938, 0.7938], c:[0.5557, 0.7938], a:[0.2967,0.4116], g:[0.1834, 0.4322], b:[0.7577, 0.8315], d:[0.1685, 0.3361], e:[0.0271, 0.0949], f:[0.9050, 0.9727]$.

Then,

$Answer-set1 = v \cup \{y:[0,0], z:[1,1], l:[0.4,0.6]\}$.

$Answer-set2 = v \cup \{y:[0.25,0.25], z:[0.75,0.75], l:[0.3,0.45]\}$.

$Answer-set3 = v \cup \{y:[0.75,0.75], z:[0.25,0.25], l:[0.1,0.15]\}$.

$Answer-set4 = v \cup \{y:[1,1], z:[0,0], l:[0,0]\}$.

\section{Application:}

Nowadays rule-based expert systems, decision support systems (DSS) are being widely used in various application domains. Artificial intelligence-based systems are being developed for helping doctors making clinical decisions. An automated Triage system can be helpful in emergency conditions in accident sites \cite{golding2008emergency,bauters2014semantics}. Apart from that, various knowledge-based clinical decision support systems (CDSS) \cite{berner2007clinical,safdari2016developing,patel2012decision,gorgulu2016use,malmir2017medical} are being developed to assist doctors in clinical decision making, where the aim is to build a computer program that could simulate human thinking. A CDSS mainly consists of three parts; 1. a knowledge base, 2. an inference engine and 3. mechanism to communicate with the user. The knowledge base captures the domain knowledge of doctors and their opinions in the form of 'IF-THEN' rules, and the inference engine deduces conclusions using this knowledge base and the specific data presented by the patient. As compared to the machine learning based CDSSs, rule-based systems is more intelligible and modular, making it easy to recognize and remove problematic rules \cite{caruana2015intelligible}. 

Some of the parameters concerned with a patient's medical condition are qualitative, which have innate imprecision. Moreover, the natural way to express doctors' knowledge is by means of using linguistic variables which express the imprecision (or vagueness). For instance, it is more natural for a doctor to say "An \textit{older} patient with \textit{severe} stomach ache has a \textit{more serious} condition than a \textit{young} patient"; rather than "A patient with age $>40$, with a stomach ache of intensity  6 or more in a scale of 0-10 is 30\% more serious than that of a patient with age $<15$". These linguistic variables and qualitative attributes can only be captured in Fuzzy Logic. Some decisions are complex and depend on too many parameters and incur some uncertainty; e.g., " A round, opacified area seen in the lungs on a chest radiograph is \textit{probably} Pneumonia" \cite{berner2007clinical}. This rule doesn't involve any imprecise attributes, rather a decision which is precise but uncertain. Hence, possibilistic or probabilistic logic may be used for capturing this uncertainty. Medical decision making is often nonmonotonic. Suppose in a scenario a patient arrives at the emergency  section with critical condition and his diagnosis requires some medical test which is time consuming or that testing facility may not be available at that instant. In that case, based on the condition of the patient, medical decision has to be done without that test result, based on other symptoms and \textit{rules of thumb}. Later, when the test result arrives, the prior decision may be found to be wrong. Hence in the inference system there must be a provision to update or revise decisions without invalidating the rules used to deduce the conclusion. Therefore, nonmonotonicity is also an inseparable part of a CDSS.

The above discussion, though focused on a specific application domain, points out the necessity of a unified \textbf{nonmonotonic} reasoning framework for handling \textbf{both} vague information and uncertain information. The approach proposed in this paper is suitable for such an application and it is expected to perform in a more intuitive way than the other proposed approaches based on solely fuzzy logic or possibilistic logic \cite{chen2013rule,rahaman2014belief,jiang2017research}. This is the subject of our future work.

\section{Conclusion:} This paper presents a semantics for unified logic program that can handle nonmonotonic reasoning with vague and uncertain information. The chosen set of truth values is the set of all sub-intervals of $[0,1]$, ordered in terms of degree of truth and degree of certainty using a preorder-based triangle. This use of preorder-based triangle instead of bilattice-based triangle distinguishes this work from any other previously proposed approach. Weighted rules are used, where rules weights are intervals depicting the degree of uncertainty. Weighted rules can be used to distinguish between propositions and dispositions (propositions having exceptions as per Zadeh), thus allowing us to perform nonmonotonic reasoning. Both classical negation and negation-as-failure are considered here. A special knowledge aggregation operator is used to take care of the interaction of positive and negative evidences for a piece of information. This operator makes the nonmonotonic reasoning more intuitive. Lastly, an iterative approach for computation of the answer set is presented here, which is influenced by the three stages of computation of classical answer sets. However, the truth-values being real numbers, to guarantee the termination of iterations become difficult. Iterations are mathematically investigated by means of difference equations, which are obtained by graphically representing rules as computation graph-like structures, namely value-propagation-graph. This analysis specifies the conditions under which the sufficient condition for convergence (for any starting value) is satisfied. The aim of this analysis is to give a glimpse of such a method of analysis. But, whether we can specify the necessary conditions for convergence or whether we can explain the convergence of programs not satisfying the sufficient condition using the value propagation graph is of further study.

\textbf{Acknowledgment:} The first author acknowledges the scholarship obtained from Department of Science and Technology, Government of India, in the form of INSPIRE Fellowship.

\bibliographystyle{spmpsci}
\bibliography{biblist1}

\section{Appendix:}

$\bullet$ \textbf{Proof of Theorem 1:} It is to prove that for any UnASP program, without any classically negated literal, there is at least one interpretation which is k-minimal supported model of the program. 

Let's assume that $P_{pos}$ is a UnASP program containing no classically negated literal. Also assume that $P_{pos}$ doesn't have supported model, i.e. for each model one of the three conditions of Definition 5 fails for some rule in $P_{pos}$.

If any rule $r_P$ in $P_{pos}$, whose head atom, say '$a$', doesn't occur in any other rule-head of $P_{pos}$, does not have any supported model, then it implies that for every interpretation $\Im$, $v_{\Im}(r_{P-Body})\notin \mathscr{T}$. The main connective of the rule body is the product t-norm $(\wedge)$, and $\wedge$ is closed over $\mathscr{T}$. So for any chosen intervals of $\mathscr{T}$ chosen as epistemic states of the literals of the rule body, their t-norm gives an interval from $\mathscr{T}$. Therefore, there has to be some $x \in \mathscr{T}$ such that $v_{\Im}(r_{P-Body}) = x$ and $\Im$ can be the supported model for rule $r_P$ by making $v_{\Im}(a) = x$.

Now suppose, there are m rules $r_1, r_2,..., r_m$ having same head $a_m$. From the above line of reasoning it can be said that, using some interpretation $\Im$ the epistemic states of the bodies of each of the rules $r_1, r_2,..,r_m$ can be obtained. Let $v_{\Im}(r_{i-Body}) = x_i$ for $1 \leq i \leq m$. Then $\Im$ would be a supported model of $r_1, r_2,..,r_m$ if $v_{\Im}(a_m) = x_1 \vee x_2 \vee ...\vee x_m$. As $\vee$ is closed over $\mathscr{T}$ there is always an interval from $\mathscr{T}$ which is equal to $x_1 \vee x_2 \vee ...\vee x_m$ that can be assigned as epistemic state of $a_m$.

Now since there isn't any classically negated literal in $P_{pos}$ there is no question of violating condition (iii) of Definition 5.

Hence our initial assumption that $P_{pos}$ doesn't have any supported model is incorrect and $P_{pos}$ must have at least one supported model. If $P_{pos}$ has a unique supported model then it is the answer set.  If more than one supported model is there, then there must be a k-least model or more than one k-minimal model. Thus, the program must have at least one answer set. (\textbf{Q.E.D})

$\bullet$ \textbf{Proof of Theorem 2:} Suppose $\Im$ is a supported model of a UnASP program $P$. For a model to be supported it has to satisfy the three conditions of supportedness as given in Definition 5. Suppose, $P$ contains following set of rules:

$r_1^a: a \stackrel{\alpha_1}{\leftarrow} B_1, \ldots, r_n^a: a \stackrel{\alpha_n}{\leftarrow} B_n$ and $r_1^{\neg a}: \neg a \stackrel{\beta_1}{\leftarrow} C_1, \ldots, r_m^{\neg a}: \neg a \stackrel{\beta_m}{\leftarrow} C_m$.

$B_1,..,B_n,C_1,..,C_m$ are body of the corresponding rules.

The corresponding rule in the transformed program $P^*$ is;

$r^{a*}: a \longleftarrow ((\alpha_1 \wedge B_1) \vee..\vee (\alpha_n \wedge B_n)) \otimes_k \neg ((\beta_1 \wedge C_1) \vee..\vee (\beta_m \wedge C_m))$.

For an interpretation $\Im$ to be a supported model of $P$;

$\Im(a) = max_k{((\alpha_1\wedge v_{\Im}(B_1)) \vee..\vee(\alpha_n\wedge v_{\Im}(B_n))), \neg ((\beta_1\wedge v_{\Im}(C_1)) \vee..\vee (\beta_m\wedge v_{\Im}(C_m)))}$

$ = [((\alpha_1\wedge v_{\Im}(B_1)) \vee..\vee(\alpha_n\wedge v_{\Im}(B_n)))] \otimes_k [((\beta_1\wedge v_{\Im}(C_1)) \vee..\vee (\beta_m\wedge v_{\Im}(C_m)))]$

$ = v_{\Im}(r^{a*}_{Body})$.

Hence, $\Im$ is also a supported model of $P^*$.\textbf{Q.E.D}

$\bullet$ \textbf{Proof of Theorem 3:} The monotonic iteration starts with the empty interpretation (partial). As iteration progresses, evaluation of some rules assign epistemic states to some atoms. Then these newly derived epistemic states are used for drawing further inferences. Once an epistemic state is assigned to some atom, that epistemic state remains unaltered throughout the MI stage (since, each atom occurs in the head of at most one rule). Hence to investigate the monotonicity of the iterations we can just take into account $\Im_{MI}^{Lit}$ at each stage of iteration.

Initially $\Im_{MI_0}^{Lit} = \phi$. As iteration progresses, more and more atoms are added to $\Im_{MI_n}$. For any $n \geq 0$, at the n$^{th}$ iteration stage, $\Im_{MI_n}^{Lit}$ is a subset of atom base $\textbf{B}_{P^*}$. From this perspective $\Gamma$ can be thought of a mapping from $2^{\textbf{B}_{P^*}}$ to $2^{\textbf{B}_{P^*}}$. Set of all subsets of $\textbf{B}_{P^*}$ forms a complete lattice under subset operation, $\subseteq$. Moreover it can be seen that $\Im_{MI_n}^{Lit}$ is monotonically increasing with $n$, i.e., for two interpretations $\Im_1$, $\Im_2$, for $\Im_1^{Lit} \subseteq \Im_2^{Lit}$, then for some consistent transformed program $P^*$;$\{\Gamma_{P*}(\Im_1)\}^{Lit} \subseteq \{\Gamma_{P*}(\Im_2)\}^{Lit}$.

Therefore, from the above line of reasoning $\Gamma$ can be viewed as a mapping from the complete lattice ($2^{\textbf{B}_{P^*}}, \subseteq$)  to itself and the operator $\Gamma$ is monotonic. 

Moreover, $\Gamma$ is continuous, i.e. for any chain $\Im_1 \subseteq \Im_2 \subseteq ... \subseteq \Im_n$, $\bigcup_{i=1..n} \Gamma(\Im_i) = \Gamma(\bigcup_{i=1..n} \Im_i)$. It is evident since $(\bigcup_{i=1..n} \Im_i) = \Im_n$ and the monotonicity of $\Gamma$ leads to $\Gamma(\Im_1) \subseteq \Gamma(\Im_2) \subseteq ... \subseteq \Gamma(\Im_n)$ and thus $\bigcup_{i=1..n} \Gamma(\Im_i) = \Gamma(\Im_n)$.

Hence from Knaster and Tarski's fixpoint theorem $\Gamma$ will have a least fixpoint and the iteration in the monotonic iteration stage would halt at this least fixpoint. \textbf{Q.E.D}

$\bullet$ \textbf{Proof of Lemma 1:} This would be proved using Principle of Mathematical Induction.

Base Case: $n = 0$.

Elements of assumption set are initiated to $[0,1]$. Hence, for every atom $a \in AS_P$ $a_0 = [a_{1_0},a_{2_0}] = [0,1]$. For any operation in the propagation path, it is evident $0 \leq a_{1_1} \leq a_{2_1} \leq 1$, i.e., $a_{1_0} \leq a_{1_1} \leq a_{2_1} \leq a_{2_0}$. 

Induction Hypothesis: Suppose for the $(m-1)^{th}$ and $m^{th}$ iterations, with $m >1$, for every atom $a \in AS_P$, 

\begin{center}

$a_{1_{(m-1)}} \leq a_{1_{m}} \leq a_{2_{m}} \leq a_{2_{(m-1)}}. .......... (i)$

\end{center}

Then it is to be proved, $\forall a \in AS_P, a_{1_m} \leq a_{1_{(m+1)}} \leq a_{2_{(m+1)}} \leq a_{2_m}$.

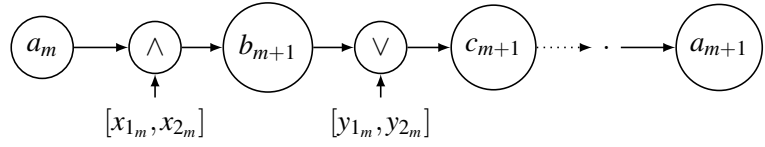
\begin{figure}
\centering
\begin {tikzpicture}[-latex ,auto ,node distance =1.0 cm and 1.5 cm ,on grid ,
semithick ,
state/.style ={ circle,
draw, text=black , minimum width = 0.5 cm}]

\node[state] (A){$a_m$};
\node[state] (Con)[right =of A]{$\wedge$};
\node (ca) [below =of Con]{$[x_{1_m}, x_{2_m}]$};
\node[state] (B)[right =of Con]{$b_{m+1}$};
\node[state] (Dis) [right =of B]{$\vee$};
\node (da) [below =of Dis]{$[y_{1_m}, y_{2_m}]$};
\node[state] (C)[right =of Dis]{$c_{m+1}$};
\node (dott) [right =of C]{$.$};
\node[state] (G)[right =of dott]{$a_{m+1}$};

\path (A) edge (Con);
\path (Con) edge (B);
\path (B) edge (Dis);
\path (ca) edge (Con);
\path (da) edge (Dis);
\path (Dis) edge (C);
\path (dott) edge (G);

\draw [dotted] (C) edge (dott);

\end{tikzpicture}
\caption{Typical example of propagation path}
\label{figure:typical}
\end{figure}

For any atom $a \in AS_P$, $a_{m+1}$ is obtained from $a_m$ by passing through the path for $a$ in the value propagation graph. Typically any such path would look like as shown in Figure \ref{figure:typical}, where, $[x_{1_m},x_{2_m}]$ and $[y_{1_m},y_{2_m}]$ (and all other conjuncts or disjuncts not shown in the figure) can be fixed intervals from $\mathscr{T}$, which remain constant for all iterations, or they can be some atoms from $As_P$ with their value at the $m^{th}$ iteration. So, in any case, generally, it can be written from the induction hypothesis 

\begin{center}

$x_{1_{(m-1)}} \leq x_{1_m} \leq x_{2_m} \leq x_{2_{(m-1)}}, \ y_{1_{(m-1)}} \leq y_{1_m} \leq y_{2_m} \leq y_{2_{(m-1)}}$....... (ii)

\end{center} 

In other words, in terms of ordering intervals based on the Bilattice-based triangle \cite {arieli2004relating}, it can be said from equations (i) and (ii) that;

\begin{center}

$a_{m-1} \leq_k a_m$ , $x_{m-1} \leq_k x_m$ and $y_{m-1} \leq_k y_m$............(iii)

\end{center}

Now, all the logical operators involved in the value propagation path of $a$, i.e., $\wedge, \vee$ and $\neg$ are monotonic with respect to $\leq_k$ \cite {arieli2004relating}. That is to say, from equation (iii) the following can be said:

\begin{center}

$b_m = a_{m-1} \wedge x_{m-1} \leq_k a_m \wedge x_m = b_{m+1};$

\end{center}

Moreover,

\begin{center}

$c_m = b_{m-1} \vee y_{m-1} \leq_k b_m \vee y_m = c_{m+1};$

\end{center}

Also, $\neg c_{m} \leq_k \neg c_{m+1}$.

Proceeding this way along the value propagation path of $a$ it is obtained that, $a_m \leq_k a_{m+1}$, i.e., $a_{1_m} \leq a_{1_{(m+1)}} \leq a_{2_{(m+1)}} \leq a_{2_m}$.

Therefore, following the principle of induction, the theorem is proved for every atom $a \in As_P$ for any $n>0$. (\textbf{Q.E.D})

$\bullet$ \textbf{Proof of Theorem 4} From Lemma 1, it can be seen that NMI iterations are monotonic with respect to the knowledge ordering $\leq_k$ of the epistemic states of the atoms in the assumption set. Since, for any $x,y \in \mathscr{T}, x \leq_k y \Rightarrow x \leq_{k_p} y$ \cite{ray2018preorder}, NMI iterations are monotonic with respect to $\leq_{k_p}$ as well;

i.e. $\Im_{NMI}^n \leq_{k_p} \Im_{NMI}^{n+1}$.

All the operations involved are continuous and since, $\otimes_k$ does not occur in the program the program is consistent. So NMI iterations are continuous as well. Hence, iteration terminates at the least fixed point.

The fixed point is unique and it is k-minimal. From the construction of $\Gamma$ on transformed program, the resulting model is also supported. Hence it is the unique answer set (following Theorem 2), that is reached by iterations. (\textbf{Q.E.D})\\

$\bullet$ \textbf{Proof of Theorem 7:} The dependency graph of $P$ being a simple cycle there is only one element (say $a$) in $As_P$ and the corresponding value propagation graph will be composed of just a single path from $a_{n-1}$ to $a_n$. Therefore the system of difference equations corresponding to the NMI iterations will be of the form:

\[
  \bar{\textbf{a}}^n =
  \begin{bmatrix}
           a_{1_n} \\
           a_{2_n}
  \end{bmatrix} =
  \begin{bmatrix}
           f_{a_1}(a_{1_{(n-1)}}, a_{2_{(n-1)}})\\
           f_{a_2}(a_{1_{(n-1)}}, a_{2_{(n-1)}})
  \end{bmatrix} = \bar{\textbf{F}}_{sc}(\bar{\textbf{a}}^{n-1}).
\]

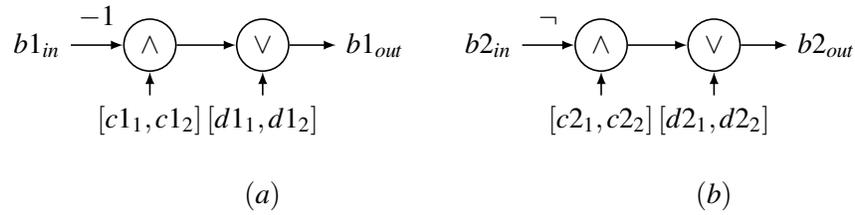
\begin{figure}
\centering
\begin {tikzpicture}[-latex ,auto ,node distance =1.0 cm and 1.5cm ,on grid ,
semithick ,
state/.style ={ circle,
draw, text=black , minimum width = 0.5 cm}]

\node (A1){$b1_{in}$};
\node[state] (C1)[right =of A1]{$\wedge$};
\node[state] (D1)[right =of C1]{$\vee$};
\node (Ao1)[right =of D1]{$b1_{out}$};
\node (nC1)[below =of C1]{$[c1_1,c1_2]$};
\node (nD1)[below =of D1]{$[d1_1,d1_2]$};
\node (ta)[below right =of nC1]{$(a)$};

\node (A2)[right =of Ao1]{$b2_{in}$};
\node[state] (C2)[right =of A2]{$\wedge$};
\node[state] (D2)[right =of C2]{$\vee$};
\node (Ao2)[right =of D2]{$b2_{out}$};
\node (nC2)[below =of C2]{$[c2_1,c2_2]$};
\node (nD2)[below =of D2]{$[d2_1,d2_2]$};
\node (tb)[below right =of nC2]{$(b)$};

\path (A1) edge node[above =0.10 cm]{$-1$} (C1);
\path (C1) edge (D1);
\path (D1) edge (Ao1);
\path (nC1) edge (C1);
\path (nD1) edge (D1);

\path (A2) edge node[above =0.10 cm]{$\neg$} (C2);
\path (C2) edge (D2);
\path (D2) edge (Ao2);
\path (nC2) edge (C2);
\path (nD2) edge (D2);

\end{tikzpicture}
\caption{Basic building Blocks of value propagation path}
\label{figure:bb1}
\end{figure}

\newcolumntype{C}{>{\centering\arraybackslash}m{22em}}
\begin{table}\sffamily
\begin{tabular}{l*2{C}@{}}
\toprule
Logical Operation & Realisation using building Blocks \\ 
\midrule

$a_n = [x,y] \wedge \text{not} \ a_{n-1}$ &

\begin {tikzpicture}[-latex ,auto ,node distance =1.0cm and 1.5cm ,on grid ,
semithick ,
state/.style ={ circle,
draw, text=black , minimum width = 0.5 cm}]

\node (A1){$a_{n-1}$};
\node[state] (C1)[right =of A1] {$\wedge$};
\node[state] (D1)[right =of C1]{$\vee$};
\node (A1o) [right =of D1]{$a_n$};
\node (nC1) [below =of C1]{$[x,y]$};
\node (nD1) [below =of D1]{$[0,0]$};

\path (A1) edge node[above =0.10cm]{$-1$} (C1);
\path (C1) edge (D1);
\path (D1) edge (A1o);
\path (nC1) edge (C1);
\path (nD1) edge (D1);

\end{tikzpicture}\\

$a_n = \neg a_{n-1}$ &

\begin {tikzpicture}[-latex ,auto ,node distance =1.0cm and 1.5cm ,on grid ,
semithick ,
state/.style ={ circle,
draw, text=black , minimum width = 0.5 cm}]

\node (A1){$a_{n-1}$};
\node[state] (C1)[right =of A1] {$\wedge$};
\node[state] (D1)[right =of C1]{$\vee$};
\node (A1o) [right =of D1]{$a_n$};
\node (nC1) [below =of C1]{$[1,1]$};
\node (nD1) [below =of D1]{$[0,0]$};

\path (A1) edge node[above =0.10cm]{$\neg$} (C1);
\path (C1) edge (D1);
\path (D1) edge (A1o);
\path (nC1) edge (C1);
\path (nD1) edge (D1);

\end{tikzpicture}\\

$a_n = ([x_1,x_2] \wedge \  a_{n-1}) \vee [y_1,y_2]$ &

\begin {tikzpicture}[-latex ,auto ,node distance =1.0cm and 1.0cm ,on grid ,
semithick ,
state/.style ={ circle,
draw, text=black , minimum width = 0.5 cm}]

\node (A1){$a_{n-1}$};
\node[state] (C1)[right =of A1] {$\wedge$};
\node[state] (D1)[right =of C1]{$\vee$};
\node (nC1) [below =of C1]{$[1,1]$};
\node (nD1) [below =of D1]{$[0,0]$};

\node[state] (C2)[right =of D1] {$\wedge$};
\node[state] (D2)[right =of C2]{$\vee$};
\node (Ao) [right =of D2]{$a_n$};
\node (nC2) [below =of C2]{$[x_1,x_2]$};
\node (nD2) [below =of D2]{$[y_1,y_2]$};

\path (A1) edge node[above =0.10cm]{$\neg$} (C1);
\path (C1) edge (D1);
\path (D1) edge node[above =0.10cm]{$\neg$}(C2);
\path (C2) edge (D2);
\path (D2) edge (Ao);
\path (nC1) edge (C1);
\path (nD1) edge (D1);
\path (nC2) edge (C2);
\path (nD2) edge (D2);

\end{tikzpicture}\\

\bottomrule 
\end{tabular}
\label{The Table}
\caption{Realisation of Logical Operation Using Building Blocks}
\end{table}

Any such value propagation path for $a$ can be constructed by composition of suitably modified building blocks shown in Figure \ref{figure:bb1}. Some examples of constructing logical operations in the value propagation path by means of the two basic building blocks are shown in Table 1. Suppose the value propagation path is constructed using blocks $b_1, b_2,..,b_n$, where, each of them are variants of the basic building blocks shown in Figure \ref{figure:bb1} and $\bar{F}_{b1}, \bar{F}_{b2}, ... , \bar{F}_{bn}$ are the iteration functions corresponding to these $b_1,..,b_n$ respectively. Then,

\begin{center}

$\bar{\textbf{F}}_{sc} = \bar{F}_{b1} \circ \bar{F}_{b2} \circ ... \circ \bar{F}_{bn}$.

\end{center} 

\textbf{Claim 1:} The iteration function $\bar{F}_{bb1}:[0,1] \times [0,1] \longrightarrow [0,1] \times [0,1]$ corresponding to the building block 1 (Figure \ref{figure:bb1} $a$) is a non-expansive mapping. $\bar{F}_{bb1}$ would be a contraction iff the path gain from $bb1_{in}$ to $bb1_{out}$, $\lVert G_{bb1} \rVert < 1$.

\textbf{Claim 2:} The iteration function $\bar{F}_{bb2}:[0,1] \times [0,1] \longrightarrow [0,1] \times [0,1]$ corresponding to the building block 2 (Figure \ref{figure:bb1} $b$) is a non-expansive mapping. $\bar{F}_{bb2}$ would be a contraction iff the path gain from $bb2_{in}$ to $bb2_{out}$, $\lVert G_{bb2} \rVert < 1$.

\textbf{Claim 3:} Composition of non-expansive mappings is a non-expansive mapping. Composition of a non-expansive mapping with a contraction mapping is a contraction. 

Given that the gain of the cycle is $<1$, there is at least one conjunct to the cycle $c \in \mathscr{T}$ that is not $[1,1]$ or at least one disjunct $d \in \mathscr{T}$ that is not $[0,0]$. So in the re-construction of the value propagation path of $a$ using building blocks, there is at least one component block to which the conjunct $c$ or the disjunct $d$ is connected; hence path gain for that block is $<1$ and that particular component gives a contraction mapping (according to claim 1 and claim 2). Therefore, following claim 3, the overall composition $\bar{F}_{b1} \circ \bar{F}_{b2} \circ ... \circ \bar{F}_{bn}$ is a contraction. Hence, the iteration function $\bar{\textbf{F}}_{sc}$, corresponding to the complete propagation path, is a contraction mapping. Moreover $\bar{\textbf{F}}_{sc}$ maps $\mathscr{T}$ into itself. Hence, from Contraction Mapping theorem it follows that, for any starting value $a_0 \in \mathscr{T}$, the iteration  $\bar{a}^k = \bar{\textbf{F}}_{sc}(\bar{\textbf{a}}^{k-1})$ will reach at the unique fixed point of $\bar{\textbf{F}}_{sc}$ in $\mathscr{T}$.

Therefore, for any starting epistemic state $a_0 \in \mathscr{T}$ NMI iteration will terminate at the unique fixed point, which is the unique answer set of the program segment under consideration.

Now it is only left to prove the claims used above.

\textbf{Proof of Claim1:} 

From Figure \ref{figure:bb1}$a$;

$bb1_{out} = [d1_1,d1_2] \vee ([c1_1,c1_2] \wedge \text{not} \ bb1_{in})$;

$[bb1_{{out}_1}, bb1_{{out}_2}] = [d1_1,d1_2] \vee ([c1_1,c1_2] \wedge \text{not} \ [bb1_{{in}_1}, bb1_{{in}_2}])$

$ \  \  \  \  \  \  \  \ \  \  \ = [d1_1,d1_2] \vee ([c1_1,c1_2] \wedge [1 - bb1_{{in}_1}, 1 - bb1_{{in}_1}])$

$ \  \  \  \  \  \  \  \ \  \  \ = [d1_1,d1_2] \vee [c1_1*(1 - bb1_{{in}_1}), c1_2*(1 - bb1_{{in}_1})]$

$ \  \  \  \  \  \  \  \ \  \  \ = [d1_1 + c1_1*(1 - bb1_{{in}_1}) - d1_1*c1_1*(1 - bb1_{{in}_1}), d1_2 + c1_2*(1 - bb1_{{in}_1}) - d1_2*c1_2*(1 - bb1_{{in}_1})]$

$ \  \  \  \  \  \  \  \ \  \  \ = [(d1_1*(1-c1_1) + c1_1) - c1_1*(1-d1_1)*bb1_{{in}_1}, (d1_2*(1-c1_2) + c1_2) - c1_2*(1-d1_2)*bb1_{{in}_1}].$\\

$bb1_{out_1} = (d1_1*(1-c1_1) + c1_1) - c1_1*(1-d1_1)*bb1_{in_1} = f_{bb1_1}(bb1_{in_1},bb1_{in_2})$;

$bb1_{out_2} = (d1_2*(1-c1_2) + c1_2) - c1_2*(1-d1_2)*bb1_{in_1} = f_{bb1_2}(bb1_{in_1},bb1_{in_2})$.\\

Thus, the mapping $\bar{F}_{bb1}: [0,1] \times [0,1] \longrightarrow [0,1] \times [0,1]$, which corresponds to the building block 1 is $\bar{F}_{bb1} = [f_{bb1_1} \  \ f_{bb1_2}]^T$

Now, for any two points $\bar{x} = (x_1,x_2), \bar{y} = (y_1,y_2) \in [0,1]^2$;

\begin{center}

$\lVert \bar{F}_{bb1}(\bar{x}) - \bar{F}_{bb1}(\bar{y}) \rVert_{\infty} =  \max\limits_{\bar{z} \in [0,1] \times [0,1]} \lVert J_{\bar{F}_{bb1}}(\bar{z}) \rVert_{\infty}  \lVert \bar{x} - \bar{y} \rVert_{\infty}$ 

\end{center}

where $J_{\bar{F}_{bb1}}(\bar{z})$ is the Jacobian matrix of $\bar{F}_{bb1}$ at $\bar{z} \in [0,1] \times [0,1]$ given by:

\[
   J_{\bar{F}_{bb1}} (\bar{z}) = 
   \begin{pmatrix}

   \frac{\partial f_{bb1_1}(\bar{z})}{\partial x_1} & \frac{\partial f_{bb1_1}(\bar{z})}{\partial x_2}\\

   \frac{\partial f_{bb1_2}(\bar{z})}{\partial x_1} & \frac{\partial f_{bb1_2}(\bar{z})}{\partial x_2}

\end{pmatrix}
\]

$\lVert J_{\bar{F}_{bb1}}(\bar{z}) \rVert_{\infty} = max (\lvert \frac{\partial f_{bb1_1}(\bar{z})}{\partial x_1} \rvert + \lvert \frac{\partial f_{bb1_1}(\bar{z})}{\partial x_2} \rvert, \lvert \frac{\partial f_{bb1_2}(\bar{z})}{\partial x_1} \rvert + \lvert \frac{\partial f_{bb1_2}(\bar{z})}{\partial x_2} \rvert)$

$= max \ (\lvert c1_1(1-d1_1) \rvert + 0, \lvert c1_2(1-d1_2) \rvert + 0)$ 

$\leq 1$ [since $0 \leq c1_1,c1_2,d1_1,d1_2 \leq 1$].

Hence, $\max\limits_{\bar{z} \in [0,1] \times [0,1]} \lVert J_{\bar{F}_{bb1}}(\bar{z}) \rVert_{\infty} \leq 1$ and thus, the mapping $\bar{F}_{bb1}$ becomes \textit{non-expansive}.

The path gain for building block 1, $\lVert G_{bb1} \rVert_{\infty} = max(c1_1(1-d1_1), c1_2(1-d1_2)) = \lVert J_{\bar{F}_{bb1}}(\bar{z}) \rVert_{\infty}$ for any $\bar{z} \in [0,1] \times [0,1]$. Therefore, if $\lVert G_{bb1} \rVert < 1$ we have,  $\max\limits_{\bar{z} \in [0,1] \times [0,1]} \lVert J_{\bar{F}_{bb1}}(\bar{z}) \rVert_{\infty} < 1$ and hence, $\lVert \bar{F}_{bb1}(\bar{x}) - \bar{F}_{bb1}(\bar{y}) \rVert < \lVert \bar{x} - \bar{y} \rVert$ for any $\bar{x}_1, \bar{x}_2 \in [0,1]^2$; i.e. $\bar{F}_{bb1}$ is a contraction mapping.

\textbf{Proof of Claim 2:}

Proceeding in the same way as in the proof of Claim 1, we get;

$bb2_{out_1} = (d2_1*(1-c2_1) + c2_1) - c2_1*(1-d2_1)*bb2_{in_2} = f_{bb2_1}(bb2_{in_1},bb2_{in_2})$;

$bb2_{out_2} = (d2_2*(1-c2_2) + c2_2) - c2_2*(1-d2_2)*bb2_{in_1} = f_{bb2_2}(bb2_{in_1},bb2_{in_2})$.\\

In the similar fashion, it can be seen;

 $\max\limits_{\bar{z} \in [0,1] \times [0,1]} \lVert J_{\bar{F}_{bb2}} (\bar{z})\rVert_{\infty} = max(c2_1(1-d2_1), c2_2(1-d2_2)) \leq 1$

i.e., $\bar{F}_{bb2}$ is a non-expansive mapping. It would become a contraction if $\lVert G_{bb1} \rVert_{\infty} < 1$.

\textbf{Proof of Claim 3:}

Suppose $F_1:[0,1]^n \longrightarrow [0,1]^n$ and $F_2:[0,1]^n \longrightarrow [0,1]^n$ are two non-expansive mappings. Now, it is to investigate whether the composition $F_2 \circ F_1$ is a non-expansive mapping.

For any $\bar{x}, \bar{y} \in [0,1]^n$, we have

$\lVert F_2 \circ F_1(\bar{x}) - F_2 \circ F_1(\bar{y}) \rVert$

$ = \lVert F_2 ( F_1(\bar{x})) - F_2 ( F_1(\bar{y})) \rVert$

$ \leq \gamma_2 \cdot \lVert F_1(\bar{x}) - F_1(\bar{y}) \rVert$ [$\gamma_2 \leq 1$ as $F_2$ is non-expansive]

$\leq \gamma_2 \cdot \gamma_1 \lVert \bar{x} - \bar{y} \rVert$ [$\gamma_1 \leq 1$ as $F_1$ is non-expansive]

$= \gamma \lVert \bar{x} - \bar{y} \rVert$ [where $\gamma = \gamma_1 \cdot \gamma_2 \leq 1$]

Thus, $F_2 \circ F_1$ is a non-expansive mapping. 

In particular, if at least one of $F_1$ and $F_2$ is contraction, then $\gamma_1 \cdot \gamma_2 < \gamma < 1$. Hence, $F_2 \circ F_1$ becomes a contraction.

This proof can be extended to composition of any number of functions by using induction. (\textbf{Q.E.D})

$\bullet$ \textbf{Proof of Lemma 2} The proof is done using mathematical induction.

\textbf{Base case:}  m = 2.

Case 1. $f_2 = a_1.a_2$ (product t-norm)

$\lvert \frac{\partial f_2}{\partial a_1}\rvert + \lvert \frac{\partial f_2}{\partial a_2}\rvert$ = $|a_1| + |a_2| \leq 2$.

Case 2. $f_2 = a_1 + a_2 - a_1.a_2$ (product t-conorm)

$\lvert \frac{\partial f_2}{\partial a_1}\rvert + \lvert \frac{\partial f_2}{\partial a_2}\rvert$ = $|1 - a_1| + |1 - a_2| \leq 2$.

since, $0 \leq a_1, a_2 \leq 1$.

Case 3. $f_2 = a_1(1-a_2)$ (product with negated variable)

$\lvert \frac{\partial f_2}{\partial a_1}\rvert + \lvert \frac{\partial f_2}{\partial a_2}\rvert$ = $|1 - a_2| + |- a_1| \leq 2$.

The same can be shown for the rest of the combinations using De Morgan's Law for negation.

\textbf{Induction hypothesis:} Say $f_n$ be a logical function of $n$ variables satisfying the aforementioned condition, i.e.

\begin{center}
$\lvert \frac{\partial f_n}{\partial a_1}\rvert + \lvert \frac{\partial f_n}{\partial a_2}\rvert + ... + \lvert \frac{\partial f_n}{\partial a_n}\rvert \leq n$.

\end{center}

Now, any logical function with $(n+1)$ logical variable can be constructed from basic logical operations on $f_n$ and $a_{n+1}$.

Case 1: $f_{n+1} = f_n.a_{n+1}$ (product t-norm)

$\lvert \frac{\partial f_{n+1}}{\partial a_1}\rvert + \lvert \frac{\partial f_{n+1}}{\partial a_2}\rvert + ... + \lvert \frac{\partial f_{n+1}}{\partial a_{n+1}}\rvert$

$ = a_{n+1}(\lvert \frac{\partial f_n}{\partial a_1}\rvert + \lvert \frac{\partial f_n}{\partial a_2}\rvert + ... + \lvert \frac{\partial f_n}{\partial a_n}\rvert) + f_n$

$\leq n.a_{n+1} + 1$

$\leq n+1$.

Case 2: $f_{n+1} = f_n + a_{n+1} - f_n.a_{n+1}$ (t-conorm)

or,  $f_{n+1} = a_{n+1} + f_n(1-a_{n+1})$

$\lvert \frac{\partial f_{n+1}}{\partial a_1}\rvert + \lvert \frac{\partial f_{n+1}}{\partial a_2}\rvert + ... + \lvert \frac{\partial f_{n+1}}{\partial a_{n+1}}\rvert$

$= (1-a_{n+1})(\lvert \frac{\partial f_n}{\partial a_1}\rvert + \lvert \frac{\partial f_n}{\partial a_2}\rvert + ... + \lvert \frac{\partial f_n}{\partial a_n}\rvert) + (1-f_n)$

$\leq  (1-a_{n+1})n + (1-f_n)$

$\leq n+1$.

Case 3: $f_{n+1} = (1-a_{n+1})f_n$ (product with negated variable)

$\lvert \frac{\partial f_{n+1}}{\partial a_1}\rvert + \lvert \frac{\partial f_{n+1}}{\partial a_2}\rvert + ... + \lvert \frac{\partial f_{n+1}}{\partial a_{n+1}}\rvert$

$= (1-a_{n+1})(\lvert \frac{\partial f_n}{\partial a_1}\rvert + \lvert \frac{\partial f_n}{\partial a_2}\rvert + ... + \lvert \frac{\partial f_n}{\partial a_n}\rvert) + f_n$

$\leq n+1$.

For rest of the combinations the proof can be derived using De Morgan's Law. Therefore, following the principle of mathematical induction the theorem is proved for any $m \geq 2$. \textbf{Q.E.D}

$\bullet$ \textbf{Proof of Theorem 8}: The path gain of the vpp for atom $a$ can be calculated by using Algorithm 1 using the constant conjuncts only. Since in the vpp of the atom $a$ no disjunctions are there, the two component functions of the iteration function corresponding to the atom $a$ will be of the form

$\bar{f}_{a} = [f_1 \ f_2]^T$, where, 

$f_1 = G_1.f_{a_1}(a_1,a_2,g_1,...,g_k)$ and $f_2 = G_2.f_{a_2}(a_1,a_2,g_1,...,g_k)$.

$f_1$ and $f_2$ comprise two rows of the Jacobian matrix of the whole SCC. The infinite norm of $\bar{f}_{a}$ is given by:

$max(\lvert \frac{\partial f_{1}}{\partial a_1}\rvert + \lvert \frac{\partial f_{1}}{\partial a_2}\rvert + ... + \lvert \frac{\partial f_{1}}{\partial g_k}\rvert, \ \lvert \frac{\partial f_{2}}{\partial a_1}\rvert + \lvert \frac{\partial f_{2}}{\partial a_2}\rvert + ... + \lvert \frac{\partial f_{2}}{\partial g_k}\rvert)$

$ = max(G_1(\lvert \frac{\partial f_{a_1}}{\partial a_1}\rvert + \lvert \frac{\partial f_{a_1}}{\partial a_2}\rvert + ... + \lvert \frac{\partial f_{a_1}}{\partial g_k}\rvert), \ G_2(\lvert \frac{\partial f_{a_2}}{\partial a_1}\rvert + \lvert \frac{\partial f_{a_2}}{\partial a_2}\rvert + ... + \lvert \frac{\partial f_{a_2}}{\partial g_k}\rvert)$

$\leq max(G_1.(k+2), G_2.(k+2))$ \  \ [following Lemma 2]

$\leq (k+2)\cdot max(G_1,G_2)$

$\leq 1$. [since path gain = $\lVert G \rVert_{\infty}$ = $max(G_1,G_2) \leq \frac{1}{k+2}$] \textbf{Q.E.D}

$\bullet$ \textbf{Proof of Theorem 9} 

From the construction of NMI iteration over the transformed programs 1 and 2 are guaranteed.

For the rest of the claims of Theorem 9 it is sufficient to show that the program $P_c$ can have no other answer sets than the ones specified in the theorem.

Suppose the contrary is true. There is another answer set $\Im$ which assigns an interval $[a_1,a_2]$ to $a$ so that $[a_1,a_2] \neq \bar{c}$ and $[a_1,a_2] \neq v_{\Im^{-\otimes}}(B_1)$.

If, $[a_1,a_2] \leq_{k_p} \bar{c}$, then $[a_1,a_2]$ cannot be a supported epistemic state for $a$, because $\otimes_k$ in the rule body can not allow an epistemic state with lower knowledge degree than $\bar{c}$ to the head of the rule $r$.

Then, suppose, $\bar{c} \leq_{k_p} [a_1,a_2]$. Thus it can be said $v_{\Im}(B_1) = [a_1,a_2]$. This effectively reduces $P_c$ to $P_c^{-\otimes_k}$, because now $\bar{c}$ has no effect on the rule head. It is claimed that $P_c^{-\otimes_k}$ gives rise to a contraction mapping and it has a unique answer set which assigns $v_{\Im^{-\otimes}}(B_1)$ to $B_1$ and hence $[a_1,a_2] \neq v_{\Im^{-\otimes}}(B_1)$ is contradictory. So it is proved the program $P_c$ can have at most two answer sets as specified in Theorem 9 and nothing else. (\textbf{Q.E.D})

\end{document}